\documentclass[10pt,twocolumn,letterpaper]{article}
\usepackage{anyfontsize}
\usepackage{cvpr}
\usepackage{amsmath}
\usepackage{enumitem}
\usepackage{algorithm}
\usepackage{algorithmic}
\usepackage{adjustbox}
\usepackage[accsupp]{axessibility}

\usepackage[dvipsnames]{xcolor}
\newcommand{\red}[1]{{\color{red}#1}}

\definecolor{cvprblue}{rgb}{0.21,0.49,0.74}
\usepackage[pagebackref,breaklinks,colorlinks,citecolor=cvprblue]{hyperref}

\def\paperID{8183}
\def\confName{CVPR}
\def\confYear{2024}

\title{Non-rigid Structure-from-Motion: Temporally-smooth Procrustean Alignment and Spatially-variant Deformation Modeling}

\author{Jiawei Shi,~~Hui Deng,~~Yuchao Dai \footnotemark[1]\\
School of Electronics and Information, Northwestern Polytechnical University \\  Shaanxi Key Laboratory of Information Acquisition and Processing\\
{\tt\small \{sjw2018, denghui986\}@mail.nwpu.edu.cn, daiyuchao@nwpu.edu.cn}
}

\begin{document}

\maketitle

{
\renewcommand{\thefootnote}{\fnsymbol{footnote}}
\footnotetext[1]{Yuchao Dai is the corresponding author.}
}
\begin{abstract}
Even though Non-rigid Structure-from-Motion (NRSfM) has been extensively studied and great progress has been made, there are still key challenges that hinder their broad real-world applications: 1) the inherent motion/rotation ambiguity requires either explicit camera motion recovery with extra constraint or complex Procrustean Alignment; 2) existing low-rank modeling of the global shape can over-penalize drastic deformations in the 3D shape sequence.
This paper proposes to resolve the above issues from a \emph{spatial-temporal modeling} perspective. 
First, we propose a novel Temporally-smooth Procrustean Alignment module that estimates 3D deforming shapes and adjusts the camera motion by aligning the 3D shape sequence consecutively.
Our new alignment module remedies the requirement of complex reference 3D shape during alignment, which is more conductive to non-isotropic deformation modeling.
Second, we propose a spatial-weighted approach to enforce the \emph{low-rank constraint adaptively} at different locations to accommodate drastic spatially-variant deformation reconstruction better.
Our modeling outperform existing low-rank based methods, and extensive experiments across different datasets validate the effectiveness of our method\footnote{Project page: \href{https://npucvr.github.io/TSM-NRSfM/}{https://npucvr.github.io/TSM-NRSfM}.}.

\end{abstract}

\section{Introduction}

\begin{figure}[t]
  \centering
  \includegraphics[width=1.0\linewidth]{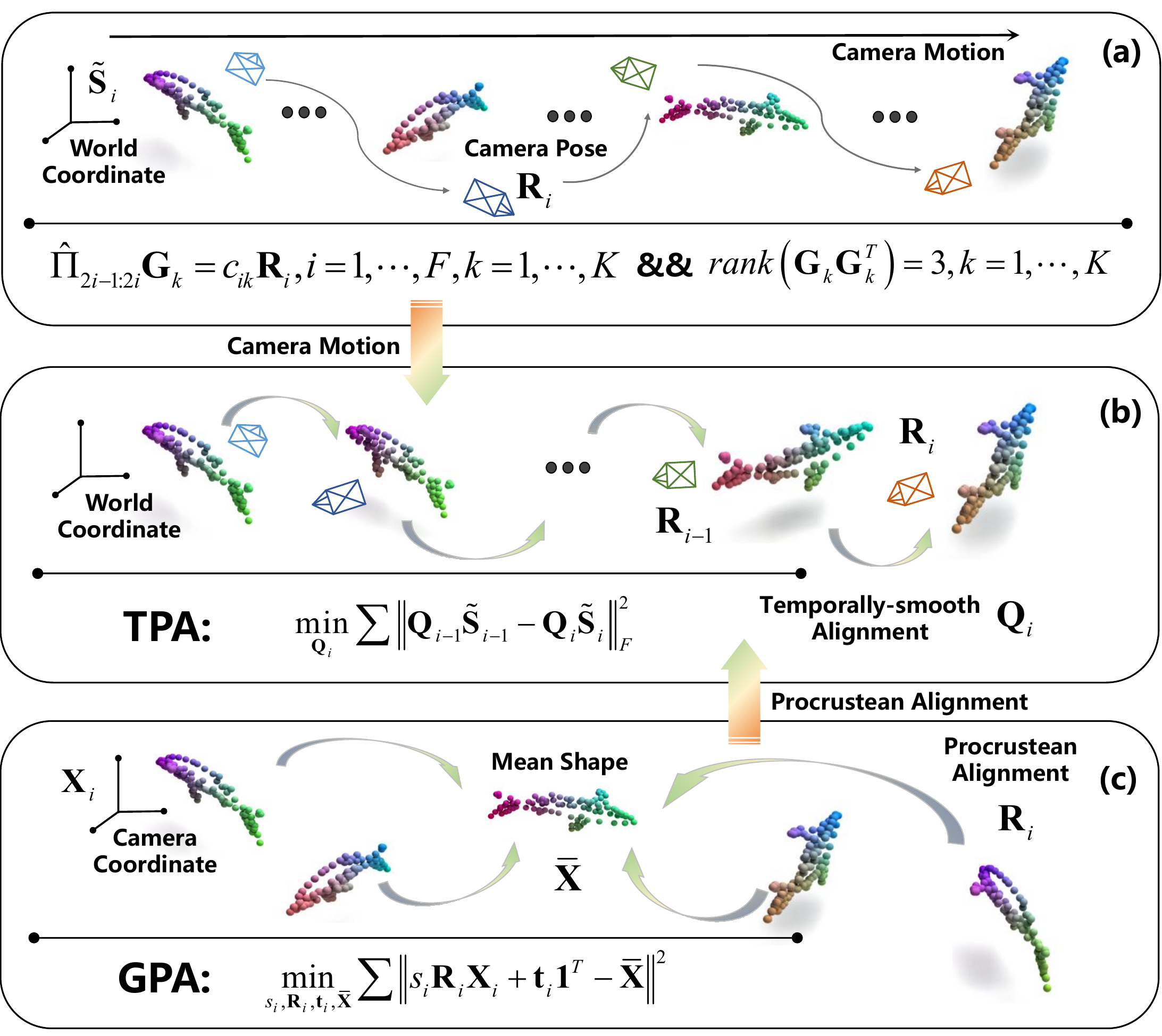}
   \caption{Overview of our proposed TPA Module. (a) Camera motion can be recovered by orthographic and rank-3 constraints within the matrix factorization framework\cite{dai2014simple}. (c) The Procrustean alignment framework uses GPA to resolve the rotation ambiguity. (b) We propose the TPA module, which aligns the 3D shapes of consecutive frames and corrects the camera motion by the temporal smoothing property.}\vspace{-1.0em}
   \label{fig:teaser}
\end{figure}

Non-rigid structure-from-motion (NRSfM) targets at jointly estimating the deforming 3D shapes and the camera motion from the 2D observation sequence, which is a classical and long-lasting geometric vision problem~\cite{bregler2000recovering}. In contrast to the recently booming deep learning based solutions ~\cite{novotny2019c3dpo, kong2020deep, park2020procrustean, zeng2022mhr, deng2022deep, Ji_2024_WACV}, the optimization based solutions do not rely on training data and can be directly applied to arbitrary classes such as human, animal, and deforming surface. Even though considerable progress has been made~\cite{dai2014simple} and a series of milestones have been achieved~\cite{lee2013procrustean, zhu2014complex,agudo2020unsupervised, kumar2022organic}, the performance of NRSfM is still far from satisfactory. The challenges mainly lie in 1) the inherent motion/rotation ambiguity requires either explicit camera motion recovery with extra constraint or complex Procrustean Alignment; 2) existing global modeling (\ie, low-rank~\cite{dai2014simple,kumar2020non,kumar2022organic}, union-of-subspace~\cite{zhu2014complex,kumar2017spatio}) of the deforming 3D shape cannot handle drastic spatially-variant 3D deformations.

Existing NRSfM statistical prior frameworks can be roughly classified into two categories:
1) explicit motion estimation-based approaches~\cite{bregler2000recovering,akhter2008nonrigid, gotardo2011computing, dai2014simple, kumar2020non, kumar2022organic} and 2) motion estimation-free approaches~\cite{lee2013procrustean, Lee_2014_CVPR, park2017procrustean}. 
The former usually employs the matrix factorization framework and orthographic constraint to solve for camera motion and 3D shapes separately.
Dai~\etal~\cite{dai2014simple} introduced the low-rank constraint into the recovery of camera motion. Kumar \etal dig into the correction matrix to improve the motion estimation results by using the smoothing prior of camera motions~\cite{kumar2020non} or the fusion of different camera motion sequences~\cite{kumar2022organic}. 
However, these camera motion estimation methods have difficulties in tackling the inherent rotation/motion ambiguity problem in NRSfM~\cite{xiao2004closed}.

The latter solves the shapes directly and remedies the requirement of motion estimation. Lee \etal~\cite{lee2013procrustean, Lee_2014_CVPR, park2017procrustean} argued that non-rigid shapes can be recovered unambiguously under the coordinate with almost no rigid relative motion. This type of framework resolves the rotation ambiguity through General Procrustean Analysis (GPA)~\cite{gower1975generalized} to achieve alignment of the 3D shapes. However, the high dependence on the alignment reference shape makes the model extremely complex~\cite{lee2013procrustean, park2017procrustean} and tends to over-penalize non-isotropic deformations.

Explicit deforming 3D sequence modeling is another difficulty in NRSfM. 
The seminal work by Dai \etal~\cite{dai2014simple} recovers the 3D shape sequence using only the low-rank constraint. Although the low-rank constraint is valid in most cases, it is insufficient to model the objects with severe spatially-variant deformation~\cite{kumar2022organic}. 
Kumar \etal~\cite{kumar2020non, kumar2022organic} improve the estimation of low-rank structures but still use global low-rank modeling. 
Some methods~\cite{grasshof2022tensor, zeng2022mhr} tried to introduce additional combinations of linear basis to model drastic deformation, but the lack of effective constraints makes the recovery of deformation basis difficult.

In this paper, we propose to resolve the above issues in camera motion and deforming shape recovery from a \emph{spatial-temporal modeling perspective}. The object of our study is the \emph{2D observation sequences}, and this is a common setup for traditional NRSfM~\cite{akhter2008nonrigid,gotardo2011computing,gotardo2011non,Lee_2014_CVPR,park2017procrustean}. Existing methods do not utilize sequence information well, which can help us deal with the above problems effectively.
First, by digging into the temporal smoothing property of non-rigid sequences, we propose a novel \emph{Temporally-smooth Procrustean Alignment (TPA)} module that corrects camera motion through 3D shape alignment.
The TPA module can reduce the effects of rotation ambiguity as GPA by exploiting the temporal similarity between consecutive 3D shapes, and it also eliminates the need for additional modeling of reference shapes as the GPA-based methods~\cite{lee2013procrustean,park2017procrustean}(\cf~\cref{fig:teaser}).
Second, to tackle the drastic spatially-variant deformations in real-world 3D shape sequences, we propose to use the spatial information obtained by analyzing the trajectory space to segment the regions with different levels of non-rigid deformation. Afterward, we propose a \emph{Spatial-Weighted Nuclear Norm (SWNN)} optimization to improve the adaptation performance of low-rank constraint for severe spatially-variant deformation.

Our main contributions are summarized as:
\begin{itemize}

    \item We propose a Temporally-smooth Procrustean Alignment (TPA) module to estimate 3D deforming shapes and adjust camera motion by aligning the 3D shape sequence consecutively. 
    The TPA module is more conductive to non-isotropic deformation modeling by remedying the requirement of mean shape as alignment reference.
    
    \item 
    We contribute a spatial-weighted approach to enforce the low-rank constraint adaptively at different locations, which can better accommodate severe spatially-variant deformation than global low-rank modeling.

    \item We develop a unified spatial-temporal modeling framework for NRSfM. Extensive experiments across datasets demonstrate the superiority of our approach. Ablation studies show the effectiveness of each algorithm module.
\end{itemize}

\section{Related Work}
\vspace{-0.2em}
\subsection{Matrix Factorization based NRSfM}
\vspace{-0.2em}

Since Bregler~\etal~\cite{bregler2000recovering} first applied the factorization framework in non-rigid reconstruction, researchers begun to shift their focus from pre-learned modeling methods~\cite{blanz1999morphable, bascle1998separability} to the optimization methods, such as using Metric Projection~\cite{paladini2009factorization, paladini2012optimal}, deformable surface modeling~\cite{salzmann2007surface,varol2009template}, and some priors, \eg, motion states~\cite{han2004reconstruction}, shape priors~\cite{del2006non,del2008factorization}, DCT trajectory basis~\cite{akhter2008nonrigid, gotardo2011non}, \etc. However, Xiao~\etal~\cite{xiao2004closed} pointed out that, unlike the rigid version of factorization framework~\cite{tomasi1992shape}, there is an inherent ambiguity in the solution process using only the orthographic constraint, which would lead to non-unique shape basis and corresponding coefficients. But Akhter~\etal~\cite{akhter2009defense} later proved that inherent ambiguity does not affect obtaining a unique result. Based on this, Dai \etal~\cite{dai2014simple} proposed a new pipeline that employs low-rank constraint on camera motion and shape estimation, combined with the solution technique of Brand~\cite{brand2005direct}, this method successfully achieves an unambiguous shape estimation using the factorization framework without introducing an overly strong prior.

Subsequent researchers have derived a number of different solution methods based on the application of low-rank constraints in the factorization framework. Kumar \etal~\cite{kumar2020non, kumar2022organic} noted that the reconstruction accuracy of 3D shapes can be improved by preserving the main algebraic features of the matrix in the nuclear norm optimization process. In addition,
~\cite{zhu2014complex, kumar2017spatio, agudo2020unsupervised} model 3D shapes using union of subspace constraint,~\cite{lee2016consensus, cha2019reconstruct} model spatial structures using consensus assumptions, and some work represent non-rigid 3D surfaces using manifolds\cite{kumar2018scalable, parashar2019local, parashar2021robust}. The factorization framework has been shown to be effective in NRSfM, but a number of issues remain. The accuracy of the unique shape obtained under the existing pipeline is compromised by the inherent ambiguity, and global low-rank modeling does not well tackle the reconstruction of severely deformed objects. And this paper intends to fill those gaps. 

\subsection{Procrustean Alignment for NRSfM} 
Parallel to the development of the factorization framework, the alignment framework chose to reduce the impact of the inherent ambiguity on reconstruction by avoiding estimating motion. Torresani \etal~\cite{torresani2008nonrigid} used Gaussian distribution as a prior to represent non-rigid shapes and proposed an Expectation Maximization (EM) solving framework. To separate rigid motion and non-rigid deformation, Lee \etal~\cite{lee2013procrustean} introduced the General Procrustean Analysis (GPA) algorithm to construct a special Procrustean Normalized Distribution (PND) that effectively represents the data properties of non-rigid shapes. Lee \etal~\cite{Lee_2014_CVPR} added the hidden Markov process to PND, strengthening the temporal dependence in the shape sequence. Park \etal~\cite{park2017procrustean} designed a more extensible regression framework based on PND and applied it to deep learning method~\cite{park2020procrustean}. 
The separation operation of the alignment framework has a significant effect on mitigating the rotation ambiguity than most of factorization methods, but on the other hand, the pure alignment framework fails to capitalize on the advantages of the factorization framework as well. Therefore, this paper aims to combine the advantages of both to obtain better results.

\section{Method}\label{section 3}
In this section, we introduce our spatial-temporal modeling solution for NRSfM. We first recap the definition of two classical models in NRSfM. Then we improve the modeling of non-rigid deformations from both temporal and spatial perspectives, \ie, temporally-smooth Procrustean alignment and spatial-weighted non-rigid deformation modeling.

\subsection{Problem Formulation and Optimization} 
NRSfM aims to recover 3D deforming shape sequence $\mathbf{S}\!=\![\mathbf{S}_{1};\cdots;\mathbf{S}_{F}]\!\in\!\mathbb{R}^{3F \times P}$ in the world coordinate and camera motion $\mathbf{R}\!=\!diag(\mathbf{R}_i)\!\in\!\mathbb{R}^{2F\times 3F}$ from 2D measurements $\mathbf{W}\!=\! [\mathbf{W}_{1};\cdots;\mathbf{W}_{F}]\!\in\!\mathbb{R}^{2F \times P}$, \ie, $\mathbf{W}=\mathbf{R}\mathbf{S}$~\cite{bregler2000recovering},
$F$ and $P$ denote the number of frames and points.
The matrix factorization model~\cite{tomasi1992shape, bregler2000recovering, dai2014simple} assumes the non-rigid 3D shapes can be expressed by the linear combination of $K$ shape basis $\mathbf{B}_{j}$, \ie, $\mathbf{S}_i\!=\!\sum_{j=1}^{K}c_{j}\mathbf{B}_{j}$, where $c_{j}$ is the shape basis coefficients. Following this assumption, $\mathbf{R}$ and $\mathbf{S}$ can be estimated by the Singular Value Decomposition (SVD) of $\mathbf{W}$ coupled with orthographic constraint.
Since $\forall~\mathbf{Q}\!\in\!SO(3), \mathbf{W}_{i}\!=\!\mathbf{R}_{i}\mathbf{Q}^{T}\mathbf{Q}\mathbf{S}_{i}$ always holds, there is an inherent rotation ambiguity in the absence of additional constraints, which leads to incorrect estimation of camera motion and 3D shapes.~\cite{dai2014simple,kumar2020non, kumar2022organic} have proposed a number of effective prior-free methods to recover camera motion and shape sequence, but these methods can not resolve the rotation ambiguity problem theoretically.

Different from the above methods, another class of methods\cite{lee2013procrustean, park2017procrustean} estimates alignment rotations by Procrustean Alignment to substitute the recovery of camera motion. The General Procrustean Alignment (GPA) model is defined as:
\begin{equation}\label{GPA}
    \mathbf{\tilde{R}}_{i}=\arg\min_{\mathbf{{R}}_{i}}\left\| \mathbf{{R}}_{i}\mathbf{S}_{i}\mathbf{T}- \mathbf{\bar{S}} \right \|, \mathbf{{R}}_{i}\!\in\!SO(3),    
\end{equation}
where $\mathbf{T}\!=\!\mathbf{I}\!-\!\frac{1}{P}\mathbf{1}\mathbf{1}^{T}$ is the translation elimination matrix, $\mathbf{\bar{S}}$ is the mean shape used as the reference shape during alignment, and $\mathbf{S}_{i}$ is the 3D shape in camera coordinate. 
Such methods effectively resolve the rotation ambiguity by separating rigid motion from non-rigid deformation. 
But for non-isotropic deformations, variations along certain directions may markedly influence the mean shape, leading to an exaggerated penalization of shapes during optimization.

In addition to the reprojection information $\mathbf{W}$ in camera coordinate, both the factorization framework and Procrustean alignment framework are looking for a canonical coordinate (\ie, world or alignment coordinate) to enforce extra regularization, such as low-rank.
We use $\mathbf{S}\in\mathbb{R}^{3F\times P}$ to denote the sequence of 3D shapes in the camera coordinate, and $\mathbf{\hat{S}}\in\mathbb{R}^{3F\times P}$ is the shape sequence in the canonical coordinate. We rethink these two types of frameworks and provide a new unified framework:
\begin{equation}
    \begin{aligned}\label{general_model}
    & \min_{\mathbf{S},\mathbf{\hat{S}},\mathbf{R}}  \mathcal{F}(\mathbf{S})+\mathcal{G}(\mathbf{\hat{S}}) +\mathcal{P}(\mathbf{R}, \Phi ), \\
    & \mathrm{ s.t. } ~ \mathbf{\hat{S}} = \mathbf{R}\mathbf{S},
    \end{aligned}
\raisetag{6\baselineskip}
\end{equation}
where $\mathcal{F}(\cdot)$ denotes the data term, such as reprojection error, and $\mathcal{G}(\cdot)$ denotes the regularization term, such as low-rank or smoothing. $\mathcal{P}(\cdot)$ represents the transformation refinement module, $\mathbf{R}$ is the transformation between the camera and canonical coordinates, $\Phi$ is the optimization parameters. $\mathbf{R}$ denotes pure 3D rotation transformation, while we have removed scale and translation from measurement matrix $\mathbf{W}$ by regularization and centralization as~\cite{dai2014simple}. 
When $\mathbf{R}$ is computed in advance and fixed during optimization, the model \eqref{general_model} degenerates to prior-free methods, such as~\cite{dai2014simple,kumar2020non,kumar2022organic}. If the optimization goal $\mathcal{P}(\cdot)$ is set to GPA \eqref{GPA} and $\Phi$ is defined as $\{ \mathbf{S},\mathbf{\bar{S}} \}$, model \eqref{general_model} degenerates into the framework in~\cite{park2017procrustean}.

\begin{figure*}
  \centering
  \begin{subfigure}{0.68\linewidth}
    \includegraphics[height=4.8cm,width=1.0\linewidth]{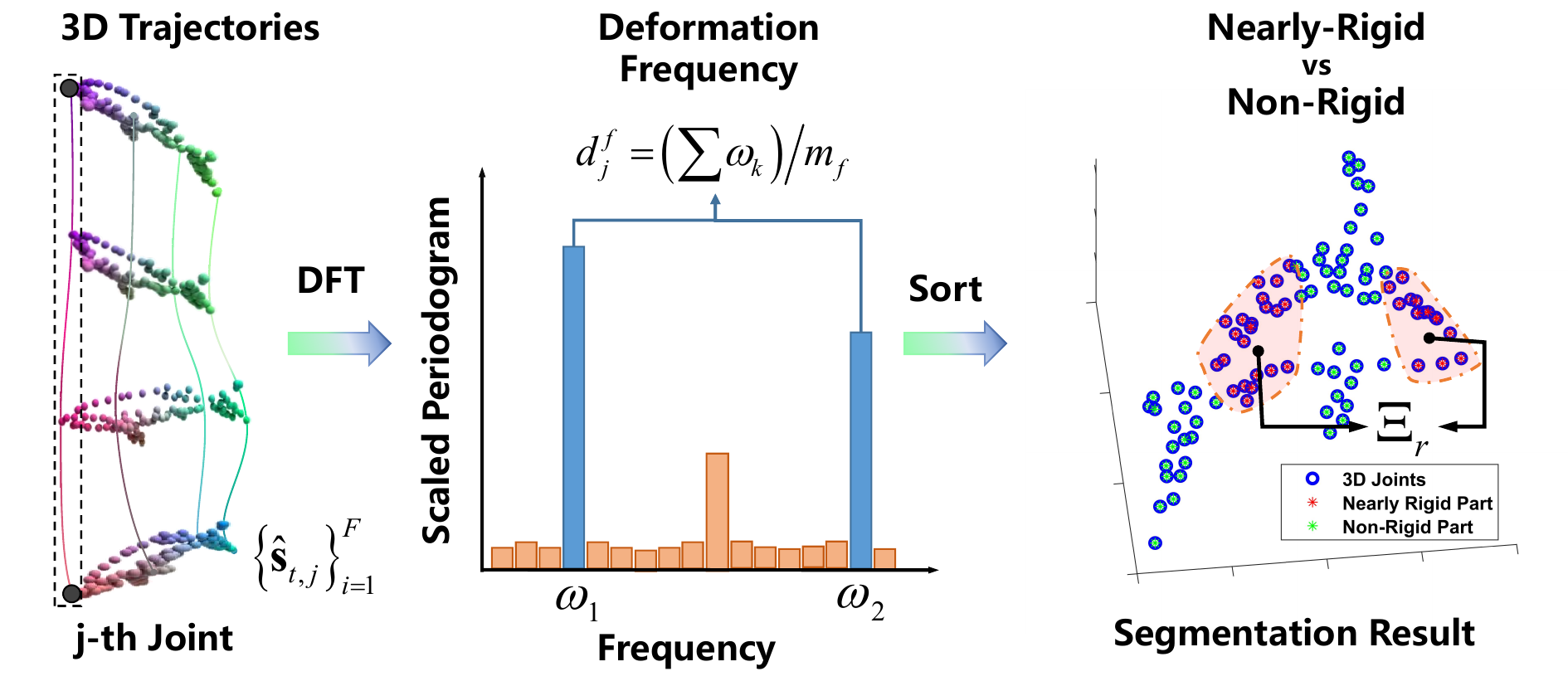}
    \caption{Process for nearly rigid region segmentation.}
    \label{fig:short-a}
  \end{subfigure}
  \hfill
  \begin{subfigure}{0.31\linewidth}
    \includegraphics[height=4.9cm,width=1.0\linewidth]{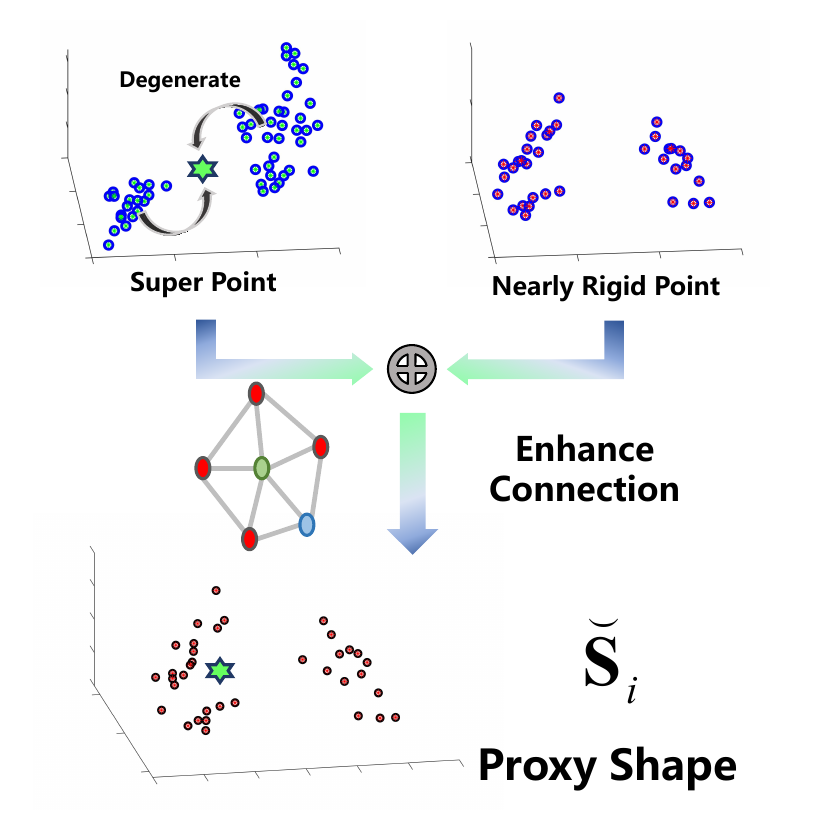}
    \caption{Construction for proxy Shape.}
    \label{fig:short-b}
  \end{subfigure}
  \caption{Overview of region segmentation and proxy shape construction. (a) We use DFT to analyze the 3D trajectories, dividing them by comparing the frequency components contained in each trajectory. The figure shows a segmentation result. (b) The geometric center of the non-rigid region is set as the super point, and it is linearly combined with nearly rigid points to construct the proxy shape.}
  \vspace{-0.4cm}
  \label{fig:short}
\end{figure*}

\subsection{Temporally-smooth Procrustean Alignment}

Under the matrix factorization framework, the 3D reconstruction results will be affected by camera motion estimation as incorrect camera motion estimation may invalidate the underlying low-rank assumption. The Procrustean alignment methods make the aligned sequence satisfy the low-rank constraint by minimizing the rigid motion between 3D shapes. Therefore, such methods can obtain improved results when they begin with good initialization.

In the real world, the nonrigid objects deform continuously and smoothly. 
Therefore, the low rank as well as other deformation regularization constraints should be satisfied in the smooth motion sequence. We propose a camera motion refinement module called \textbf{Temporally-smooth Procrustean Alignment (TPA)}, which aligns the shape sequence under the smoothness constraint as follows:
\vspace{-0.2em}
\begin{equation}\label{TPA_formula}
    \mathcal{L}_{tpa}\left ( \mathbf{R}, \mathbf{S} \right ) = \frac{1}{2}\sum_{i=1}^{F-1}\left \| {\mathbf{R}_{i}\mathbf{{S}}_{i} - \mathbf{R}_{i+1}\mathbf{{S}}_{i+1}} \right \|_{F}^{2},
\vspace{-0.2em}
\end{equation}
where $\mathbf{R} = diag\left ( \mathbf{R}_{i} \right), \mathbf{R}_{i} \in SO(3)$. The TPA module \eqref{TPA_formula} estimates alignment matrix $\mathbf{R}$ by the principle of minimizing the difference between the 3D shapes of consecutive frames after rotation transformation, as in~\cref{fig:teaser} (b). Different from GPA \eqref{GPA}, TPA does not need to estimate the mean shape of the sequence, which greatly simplifies the optimization and avoids over-penalizing non-isotropic deformations. In addition, we have experimentally verified that sequences aligned by the TPA module can have more similar low-rank and smoothing properties to the real sequences compared with GPA, see~\cref{ablation_sec} and~\cref{fig:spa_exp}.

Under the NRSfM problem setting, the TPA module can gradually put the shape sequence in a smooth state by adjusting the camera motion. 
Assume that the estimated camera poses are $\mathbf{\tilde{R}}_{i},i=1,\cdots,F$ and define the initialization of the alignment transformation as $\mathbf{R}_{pi}=\mathbf{\tilde{R}}_{i}^{T}$. We denote $\mathbf{R}_{pi}\mathbf{S}_{i}$ as $\mathbf{\tilde{S}}_{i}$, where $\mathbf{S}_{i}$ is in the camera coordinate. Then we use the TPA module to estimate the correction rotation $\mathbf{Q}=diag(\mathbf{Q}_{i})$ as follows:
\vspace{-0.3em}
\begin{equation}
    \begin{aligned}\label{TPA_module}
     \mathcal{L}^{Q}_{tpa}\left ( \mathbf{Q}, \mathbf{\tilde{S}} \right ) & = \frac{1}{2}\sum_{i=1}^{F-1}\left \| {\mathbf{Q}_{i}\mathbf{\tilde{S}}_{i} - \mathbf{Q}_{i+1}\mathbf{\tilde{S}}_{i+1}} \right \|_{F}^{2}  
     \\
    & = \frac{1}{2}\sum_{i=1}^{F-1}\sum_{j=1}^{P}\left \| {\mathbf{Q}_{i}\mathbf{\tilde{s}}_{i,j} - \mathbf{Q}_{i+1}\mathbf{\tilde{s}}_{i+1,j}}\right \|_{2}^{2},
\end{aligned}
\raisetag{6\baselineskip}
\vspace{-0.3em}
\end{equation}
where $\mathbf{\tilde{s}}_{i,j} \in \mathbb{R}^{3 \times 1}$ is the $j$-th column of $\mathbf{\tilde{S}}_{i}$. We take the Lie algebra $\phi_{i} \in so(3)$ of $\mathbf{Q}_{i}$ as the optimization variable, and the derivative of the optimization objective $\mathcal{L}^{Q}_{tpa}$ \eqref{TPA_module} with respect to $\phi_{i}$ can be derived as:
\begin{equation}\label{derivative_of_tpa}
\vspace{-0.6em}
    \frac{\partial \mathcal{L}^{Q}_{tpa}}{\partial \phi_{i}} = \sum_{j=1}^{P}{\left [ \left ( \mathbf{Q}_{i}\mathbf{\tilde{s}}_{i,j} \right )^{\wedge } \right ]^{T}}\left ( r_{i,j}^{(1)} - r_{i,j}^{(0)} \right ),
\vspace{-0.4em}
\end{equation}
where $r_{i,j}^{(m)} = \mathbf{Q}_{i-m}\mathbf{\tilde{s}}_{i-m,j} - \mathbf{Q}_{i-m+1}\mathbf{\tilde{s}}_{i-m+1,j}$, and $\left ( \mathbf{a} \right )^{\wedge}$ represents the skew-symmetric matrix of vector $\mathbf{a}$. Once the gradient is calculated, we can use traditional numerical optimization algorithms to solve model \eqref{TPA_module}, \eg, BFGS, LMF, \etc~\cite{nocedal2006numerical}, see more derivation details and discussions in the Supplementary Material (Supp.)~Sec~1. 

TPA largely separates the rigid motion between consecutive frames in the sequence $\mathbf{\tilde{S}}$ and makes it gradually converge to a smooth state. After reducing the interference of rigid motion, we can better utilize the low-rank constraint to recover non-rigid deformation of the aligned sequence $\mathbf{Q}\mathbf{\tilde{S}}$. 
In the following, we introduce the shape estimation method and unify them afterward.

\subsection{Spatial-Weighted Non-rigid Structure Model}

\subsubsection{Spatial-Weighted Nuclear Norm}
In real-world scenarios, low-rank regularization can handle many non-rigid deformation situations but always over-penalize drastic deformations. In the absence of priors to reconstruct the deformation, we can give higher degrees of freedom for shape recovery by relaxing the low-rank constraint.
An object whose overall or local deformation is very drastic can lead to its 3D shape sequence $\mathbf{\hat{S}}$ in the world coordinate not satisfying the low-rank constraint. But we can transform its spatial structure by linearly combining the trajectories of $\mathbf{\hat{S}}$ to degenerate the object to a low-rank state. Given the affinity matrix $\mathbf{\Lambda}\!\in\!\mathbb{R}^{P \times P}$, we define the sequence of \textbf{proxy shape} as $\mathbf{\breve{S}}\!=\!\mathbf{\hat{S}}\mathbf{\Lambda}$. We can then utilize low-rank constraint to recover the proxy shapes, which in turn reduces the strength of the constraint in the region of severe deformation to better recover $\mathbf{\hat{S}}$.

We use the nuclear norm $\left \|\cdot \right \|_{\ast}$ with the rearrangement operator $g(\cdot)$~\cite{dai2014simple} to build the low-rank constraint, and denote the reshuffled sequence as $\mathbf{\hat{S}}^{\sharp}\!=\!g(\mathbf{\hat{S}})\!\in\!\mathbb{R}^{F\times 3P}$. 
Different from existing methods, we propose a \textbf{Spatial-Weighted Nuclear Norm (SWNN)} optimization model that imposes low-rank constraint on the reshuffle of the proxy shapes $\mathbf{\breve{S}}^{\sharp}\!=\!g(\mathbf{\hat{S}}\mathbf{\Lambda})$ instead of $\mathbf{\hat{S}}^{\sharp}$, so the new low-rank optimization model can be written as:
\begin{align}\label{Spatial_weighted_shrink}
    \min_{\mathbf{\hat{S}}}{ \| \mathbf{\breve{S}}^{\sharp}  \|_{\ast}}.
\end{align}
The key to the model \eqref{Spatial_weighted_shrink} is how to design the spatial weight matrix $\mathbf{\Lambda}$. Next, we present a method of constructing proxy shapes $\mathbf{\breve{S}}_{i}, i=1,\cdots,F$ to relax the low-rank constraint on severe deformation regions. We first introduce an algorithm for segmenting objects according to the level of deformation in different regions. Based on this, a \textbf{kernel based} method for the construction of proxy shapes is proposed.

\subsubsection{Nearly Rigid Region Segmentation}
In reality, not all regions of non-rigid objects undergo the same deformations, but are spatially-variant. For example, a talking mouth on a human face will undergo more severe deformation than the nose. Thus different regions should fit the low-rank regularization differently. In this section, we introduce the partition algorithm for different regions of non-rigid objects, see~\cref{fig:short-a} for the process.

We define $\mathcal{T}_{j}\!=\!\{ \mathbf{\hat{s}}_{t,j}\!\in\!\mathbb{R}^{3 \times 1} \}_{t=1}^{F}$ as the $j$-th column of shape sequence $\mathbf{\hat{S}}$, which represents a 3D trajectory. The x-y-z coordinates of a 3D trajectory can be viewed as time domain signals, and can be transformed into the frequency domain by Discrete Fourier Transform (DFT):
\begin{equation}
    \resizebox{0.42\textwidth}{!}{$ d(k/F)=F^{-1/2}\sum_{t=1}^{F}\mathbf{\hat{s}}_{t,j}e^{-2\pi itk/F},k=0,\cdots,F-1 $},
\end{equation}
we define $\omega_k\!=\!k/F$, which symbolize the different frequency components.
Intuitively, when the time domain signal changes drastically, its corresponding frequency domain signal will contain more high-frequency components. We can compare the severity of different time domain signal changes by calculating the intensity of different frequency components of the frequency domain signal. Here we use the \textbf{scaled periodogram} to calculate the intensity of the signal in the frequency domain. The formula is as follows:
\vspace{-0.2em}
\begin{equation}
    P_{g}(\omega_{k})=\frac{4}{F}\left | d(\omega_{k}) \right |^{2},k=0,\cdots,F-1.
\vspace{-0.2em}
\end{equation}
Then, we compare $\left \{ P_{g}(\omega_{k}) \right \}_{k}$ and filter out the frequency components $\omega_{k}$ corresponding to the first $m_{f}(=\!2)$ maxima. Their average value $d^{f}_{j} = \sum{\omega_{k}}/m_{f}$ is noted as the \textbf{deformation frequency} of the trajectory. We sort all the points according to the magnitude of the deformation frequencies $\{d^{f}_{j}\}_{j=1}^{P}$ from low to high, defining the points with low frequency of $\alpha_{r}\!\times\!100\%$ as \textbf{nearly rigid points}, and denote the set of their subscripts as $\mathbb{A}_{r}$. Here, $\alpha_{r}\!\in\!\left[ 0,1 \right]$ is a hyperparameter that depends on the deformation properties.

In this way, the whole object is divided into a nearly rigid region and a non-rigid region, as the example shown in~\cref{fig:short-a}. 
We summarize the operation above as:
\begin{equation}\label{alg:nrrs}
    \mathbb{A}_{r} =\mathcal{A}_{nrs}\left ( \mathbf{\hat{S}},\alpha_{r},m_{f} \right ).
\end{equation}

\subsubsection{Kernel Based Proxy Shape}

After completing the segmentation, we compute the proxy shape $\mathbf{\breve{S}}_{i}$ of $\mathbf{\hat{S}}_{i}$. We model the weight matrix $\mathbf{\Lambda}$ by the kernel function $k(\mathcal{T}_{i},\mathcal{T}_{j})$, where $\mathcal{T}_{j}$ is the 3D trajectory of the $j$-th point. The weight matrix can be expressed as follows:
\begin{equation}\label{proxy_shape}
    \mathbf{\Lambda} = \begin{bmatrix}
\left \langle \phi(\mathcal{T}_{1}),\phi (\mathcal{T}_{1}) \right \rangle & \cdots  & \left \langle \phi(\mathcal{T}_{1}),\phi (\mathcal{T}_{P}) \right \rangle \\ 
 \vdots & \ddots  & \vdots \\ 
 \left \langle \phi(\mathcal{T}_{P}),\phi (\mathcal{T}_{1}) \right \rangle& \cdots & \left \langle \phi(\mathcal{T}_{P}),\phi (\mathcal{T}_{P}) \right \rangle
\end{bmatrix},
\end{equation}
where $\phi(\cdot)$ represents a feature map, $ \left \langle \cdot, \cdot \right \rangle$ is the vector inner product. We use the inner product to compute the correlation between the vectors after the feature mapping as the combination weights between different points in the space. In order to increase the difference between the low-rank constraint effect on the nearly rigid and non-rigid regions, we set up the feature mapping $\phi(\cdot)$ in the following form:
\begin{equation}
    \phi(\mathcal{T}_{i})=\left\{\begin{matrix}
\sqrt{1-\delta_{r}^{2}}\mathbf{1}_{i}+\delta_{r}\mathbf{1}_{P+1},&i\in\mathbb{A}_{r}\\ 
\delta_{nr}\mathbf{1}_{P+1},&i\notin \mathbb{A}_{r}
\end{matrix}\right.
\end{equation}
where $\mathbf{1}_{i}$ is a $0$-$1$ vector in $P\!+\!1$ dimensions, taking $1$ only in the $i$-th dimension. $\delta_{r}$ is a constant, and $\delta_{nr}$ is set to $1/\sqrt{(1\!-\!\alpha_{r})P}$.
Here we provide a brief discussion of model \eqref{proxy_shape}. When $\alpha_{r}\!=\!1$, \ie, all points are in set $\mathbb{A}_{r}$, $\mathbf{\breve{S}}_{i}$ is just a shrink state of the original shape. If $\alpha_{r}\!<\!1$, all points in the non-rigid region degenerate into the same \textbf{super point} (as shown in~\cref{fig:short-b}). The super point prevents the low-rank constraint from acting directly on each point in the non-rigid region, which can help reduce the intensity of the low-rank constraint on the non-rigid region. In addition, the proxy shape calculation model \eqref{proxy_shape} ensures that the super point and nearly rigid points remain a unified whole\footnote{Refer to more analysis on proxy shape and its role in the Supp.~Sec~3.}.

\subsection{Complete Model and Solution}
In this section, we integrate the complete model with the unified framework \eqref{general_model} and perform the solution. For the data term $\mathcal{F}$, we use the reprojection constraint as the optimization target.
The SWNN model \eqref{Spatial_weighted_shrink} is set to the regularization term $\mathcal{G}$ and constrains the shape sequence under canonical coordinate. 
The link between the camera coordinate and the canonical coordinate is constructed through the TPA module \eqref{TPA_module}. The complete model is as follows:
\begin{equation}
 \resizebox{0.45\textwidth}{!}{$
\begin{aligned}\label{whole_model1}
     \min_{\mathbf{S},\mathbf{Q}}~ & \frac{\mu_{1}}{2}\left \| \mathbf{W}\!-\! \mathbf{\Pi}\mathbf{S} \right \|^{2}_{F} + \mu_{2}\left \| \mathbf{\mathbf{\breve{S}}^{\sharp}} \right \|_{*}+  \frac{\mu_{3}}{2}\sum_{i=1}^{F-1}\left \| \mathbf{Q}_{i}\mathbf{\tilde{S}}_{i}\!-\!\mathbf{Q}_{i+1}\mathbf{\tilde{S}}_{i+1} \right \|_{F}^{2} \\
    s.t. & \left\{\begin{array}{l}
\mathbf{\breve{S}}^{\sharp} = g(\mathbf{\hat{S}}\mathbf{\Lambda} )\\
\mathbf{\hat{S}}_{i}=\mathbf{Q}_{i}\mathbf{\tilde{S}}_{i},i=1,\cdots,F\\
\mathbf{\tilde{S}}_{i}=\mathbf{R}_{pi}\mathbf{{S}}_{i},i=1,\cdots,F
\end{array}\right.
\end{aligned}$}
 \raisetag{6\baselineskip}
\end{equation}
where $\mathbf{\Pi}=diag(\Pi_{i}), \Pi_{i} \in \mathbb{R}^{2\times 3}$ is the camera projection matrix, and we use orthogonal projection as~\cite{dai2014simple, kumar2020non}. $\mathbf{S} \in \mathbb{R}^{3F \times P}$ is the 3D shape sequence under the camera coordinate. Then we improve the initialized camera motion $\mathbf{R}_{p}$ by estimating $\mathbf{Q}_{i}\in SO(3)$ and compute the 3D shape sequence $\mathbf{\hat{S}} \in \mathbb{R}^{3F \times P}$ in the canonical coordinate. Finally, we introduce the kernel based weight matrix $\mathbf{\Lambda}$ to construct proxy shapes that more satisfy the low-rank constraint.

 \begin{table*}[t]
\centering
\caption{\centering 3D reconstruction errors on MoCap dataset. Our method shows advantages over many matrix factorization methods and Procrustean alignment methods. The second-best results are underlined, and the shape basis dimension $K_{s}$ is shown in brackets.}
\vspace{-0.2cm}
\label{table_mocap}
\resizebox{1.0\linewidth}{!}{
\begin{tabular}{c|c|c|c|c|c|c|c|c|c|c|c}
\hline
Data    & CSF1~\cite{gotardo2011computing}  & CSF2~\cite{gotardo2011non}  & KSTA~\cite{gotardo2011kernel}  & PND~\cite{lee2013procrustean} & PMP~\cite{Lee_2014_CVPR} & CNS~\cite{lee2016consensus} & PR~\cite{park2017procrustean}    & BMM~\cite{dai2014simple}   & R-BMM~\cite{kumar2020non} & OPM ~\cite{kumar2022organic}  & Ours     \\ \hline
Drink   & 0.0223 & 0.0223          & 0.0156 & 0.0037 & \textbf{0.0018} & 0.0431          & 0.0063 & 0.0152 & 0.0119 & 0.0071          & \underline{0.0031}(13)    \\
Pickup  & 0.2301 & 0.2277          & 0.2322 & 0.0372    & {0.0127} & 0.1281          & 0.0157 & 0.0315 & 0.0198 & 0.0152          & \textbf{0.0126}(12) \\
Yoga    & 0.1467 & 0.1464          & 0.1476 & 0.0140   & 0.0128  & 0.1845          & 0.0175 & 0.0225 & 0.0129 & 0.0122          & \textbf{0.0109}(10) \\
Stretch & 0.0710 & 0.0685          & 0.0674 & 0.0156  & 0.0124  & 0.0939          & 0.0156 & 0.0247 & 0.0144 & 0.0124          & \textbf{0.0114}(12) \\
Dance   & 0.2705 & 0.1983          & 0.2504 & 0.1454  & 0.1278  & \textbf{0.0759}          & 0.1266 & 0.1445 & 0.1491 & 0.1209         & \underline{0.0921}(13)    \\
Face    & 0.0363 & 0.0314          & 0.0339 & 0.0165 & 0.0166  & 0.0248          & 0.0164 & 0.0206 & 0.0179 & 0.0145 & \textbf{0.0144}(5)          \\
Walking & 0.1893 & 0.1035          & 0.1029 & 0.0465  & 0.0424  & \textbf{0.0396} & 0.0544 & 0.0908 & 0.0882 & 0.0816          & 0.0710(4)          \\
Shark   & \textbf{0.0081} & 0.0444 & 0.0160 & 0.0135  & 0.0099  & 0.0832          & 0.0272 & 0.2311 & 0.0551 & 0.0550          & 0.0258(6)          \\ \hline
\end{tabular}}
\vspace{-1.0em}
\end{table*}

\begin{table}[t]
\centering
\caption{Reconstruction error comparison with state-of-the-art on NRSfM Challenge dataset. We report the results in millimeters.}
\vspace{-0.2cm}
\label{table_challenge}
\resizebox{1.0\linewidth}{!}{
\begin{tabular}{c|c|c|c|c|c|c|c}
    \hline
    Data  & {\small CSF2~\cite{gotardo2011non}}   & {\small BMM~\cite{dai2014simple}}  &{\small R-BMM~\cite{kumar2020non}} &{\small AOW~\cite{iglesias2020accurate}}  & {\small BP~\cite{ornhag2021bilinear}}   & {\small OPM~\cite{kumar2022organic}}  & Ours  \\ \hline
    Articul. & 11.52 & 18.49 & 16.00 & 15.03 & 16.10 & {12.18} & \textbf{10.69} \\ %
    Balloon  & 10.14 & 10.39 & 7.84  & 8.05  & 8.29  & \textbf{6.29}  & \underline{7.28}  \\ %
    Paper    & 9.72 & 8.94  & 10.69 & 10.45 & \textbf{6.70}  & 8.86  & \underline{7.91}  \\ %
    Stretch  & 8.65 & 10.02 & 7.53  & 9.01  & 7.66  & {6.36}  & \textbf{5.43}  \\ %
    Tearing  & 12.04 & 14.23 & 16.34 & 16.20 & 11.26 & {10.91} & \textbf{10.77} \\ \hline
    \end{tabular}
}
\end{table}

 We optimize the model \eqref{whole_model1} by ADMM~\cite{boyd2011distributed}, and the Lagrange multiplier of the whole model is as follows:
 \vspace{-0.2em}
 \begin{equation}
 \resizebox{0.42\textwidth}{!}{$
 \begin{aligned}\label{admm_solution}
     \min_{\Omega} ~ \mathcal{L} & =  \frac{\mu_{1}}{2}\left \| \mathbf{W}\!-\!\mathbf{\Pi}\mathbf{S} \right \|^{2}_{F}  + \mu_{2}\left \| \mathbf{\breve{S}}^{\sharp} \right \|_{*} + \frac{\mu_{3}}{2}\sum_{i=1}^{F-1}\left \| \mathbf{Q}_{i}\mathbf{\tilde{S}}_{i}\!-\!\mathbf{Q}_{i+1}\mathbf{\tilde{S}}_{i+1} \right \|_{F}^{2} \\
     & + \frac{\beta}{2}\left \| \mathbf{\breve{S}}^{\sharp} - g(\mathbf{\hat{S}}\mathbf{\Lambda})  \right \|_{F}^{2}+\left \langle \mathbf{Y}_{1}, \mathbf{\breve{S}}^{\sharp} - g(\mathbf{\hat{S}}\mathbf{\Lambda}) \right \rangle \\
     & + \frac{\beta}{2}\left \| \mathbf{\hat{S}} - \mathbf{Q}\mathbf{\tilde{S}} \right \|_{F}^{2}+\left \langle \mathbf{Y}_{2},\mathbf{\hat{S}} - \mathbf{Q}\mathbf{\tilde{S}} \right \rangle \\
     & + \frac{\beta}{2}\left \| \mathbf{\tilde{S}} - \mathbf{R}_{p}\mathbf{{S}} \right \|_{F}^{2}+\left \langle \mathbf{Y}_{3}, \mathbf{\tilde{S}} - \mathbf{R}_{p}\mathbf{{S}} \right \rangle 
 \end{aligned}$}
 \raisetag{6\baselineskip}
 \vspace{-0.2em}
\end{equation}
where $\Omega =  \{\mathbf{S,\tilde{S},\hat{S},\breve{S}^{\sharp},Q} \}$ denotes the variables to be updated, $\{\mathbf{Y}_{n}\}_{n=1}^{3}$ are Lagrange multipliers. For the update of $\mathbf{\breve{S}}^{\sharp}$, we use the rectification algorithm in~\cite{kumar2020non}. The difference is that when calculating the weights $\Theta$, we truncate the singular values according to the shape basis dimension $K_{s}$ and normalize the weights. In real scenes, the captured images are often obscured and it is difficult to observe all the keypoints in each frame. Our proposed framework can handle the problem of missing points by simply adding visible information to the data term.
More details and formulas are provided in the {Supp.~Sec~2}.

\begin{algorithm}[t]
	\renewcommand{\algorithmicrequire}{\textbf{Input:}}
	\renewcommand{\algorithmicensure}{\textbf{Output:}}
	\caption{Overview of the Optimization Algorithm}
	\label{alg1}
	\begin{algorithmic}[1]
		\STATE \textbf{Input:} Initialize $\{ \mathbf{S},\mathbf{\tilde{S}},\mathbf{\hat{S}},\mathbf{\breve{S}^{\sharp}} \}$ via Pseudo inverse shape in~\cite{dai2014simple}, $\mathbf{R}_{p}$, $\mathbf{Q}=\mathbf{I}$, $\mathbf{\Lambda}=\mathbf{I}$, $\beta=1e^{-4}$ and $\epsilon=1e^{-6}$
		\REPEAT
		\STATE Updating variables $\{ \mathbf{\breve{S}^{\sharp}},\mathbf{\hat{S}},\mathbf{\tilde{S}},\mathbf{S}\}$ alternately according to model \eqref{admm_solution} yields $\{\mathbf{\breve{S}}{\_}^{\sharp},\mathbf{\hat{S}}{\_},\mathbf{\tilde{S}}{\_},\mathbf{S}{\_}\}$
		\UNTIL $\left \| \mathbf{S}{\_} - \mathbf{S} \right \|_{\infty}<\epsilon$
        \STATE \textbf{Initialization:} Calculate $\mathbb{A}_{r}$ by~\cref{alg:nrrs}, build weight matrix $\mathbf{\Lambda}$ by \eqref{proxy_shape}, reset $\beta = \beta_{d}$
        \REPEAT
        \STATE Updating variables $\Omega$ alternately according to model \eqref{admm_solution} yields $\{ \mathbf{\breve{S}}{\_}^{\sharp},\mathbf{\hat{S}}{\_},\mathbf{\tilde{S}}{\_},\mathbf{S}{\_},\mathbf{Q}{\_}\}$
        \UNTIL $\left \| \mathbf{S}{\_} - \mathbf{S} \right \|_{\infty}<\epsilon$
		\ENSURE  $\mathbf{S},\mathbf{Q}$
	\end{algorithmic}  
 
\end{algorithm}

\section{Experiments}

\subsection{Implementation Details and Evaluation Metric}
\textbf{Implementation Details.} The parameter settings in the ADMM optimization algorithm are the same as in the Organic Priors Method (OPM)~\cite{kumar2022organic}. The model \eqref{whole_model1} is a non-convex optimization that requires the initialization of camera motion and 3D shapes. We use the camera motion estimation algorithm in BMM~\cite{dai2014simple} to initialize $\mathbf{R}_{p}$. To build the weight matrix $\mathbf{\Lambda}$, a good initialization of the shape sequence $\mathbf{\hat{S}}$ is needed to calculate the segmentation of the non-rigid region. Since our model is a unified framework, there is no need to use other methods, which can be accomplished using only model \eqref{whole_model1}. As shown in~\cref{alg1}, we first fix the correction rotation $\mathbf{Q}$ and set $\mathbf{\Lambda}$ to the Identity matrix. After convergence, the weight matrix is calculated and all parameters $\Omega$ are well initialized. In addition, $\beta_{d}$ in~\cref{alg1} is generally $1e^{-2}$ or $1e^{0}$, and $\Psi=\{\mu_{1}, \mu_{2}, \mu_{3}, \alpha_{r}, \delta_{r}, K_{s}\}$ is adjusted to the dataset (see {Supp.~Sec~4} for more settings).

\noindent \textbf{Evaluation Metric.} We follow the setup in~\cite{kumar2022organic} using the mean normalized 3D reconstruction error metric to evaluate the shape reconstruction results on the motion capture benchmark (MoCap), semi-dense, and H3WB dataset. The metric is defined as $e_{3d}\!=\!\frac{1}{F}\sum_{i=1}^{F}\left \| \mathbf{S}_{i}^{est}\!-\!\mathbf{S}_{i}^{gt} \right \|_{F}/\left \| \mathbf{S}_{i}^{gt} \right \|_{F}$ and $\mathbf{S}_{i}^{est}$, $\mathbf{S}_{i}^{gt}$ denote the estimated 3D shape and the corresponding ground-truth (GT) value respectively. We remove the global ambiguity~\cite{akhter2008nonrigid, jensen2021benchmark} as in~\cite{kumar2022organic} before computing the 3D reconstruction error. To evaluate our approach on the NRSfM benchmark dataset~\cite{jensen2021benchmark}, we use the officially supplied metric script.

\subsection{Datasets and Results}
\begin{figure}
  \centering
  \begin{subfigure}{0.98\linewidth}
    \includegraphics[width=1.0\linewidth]{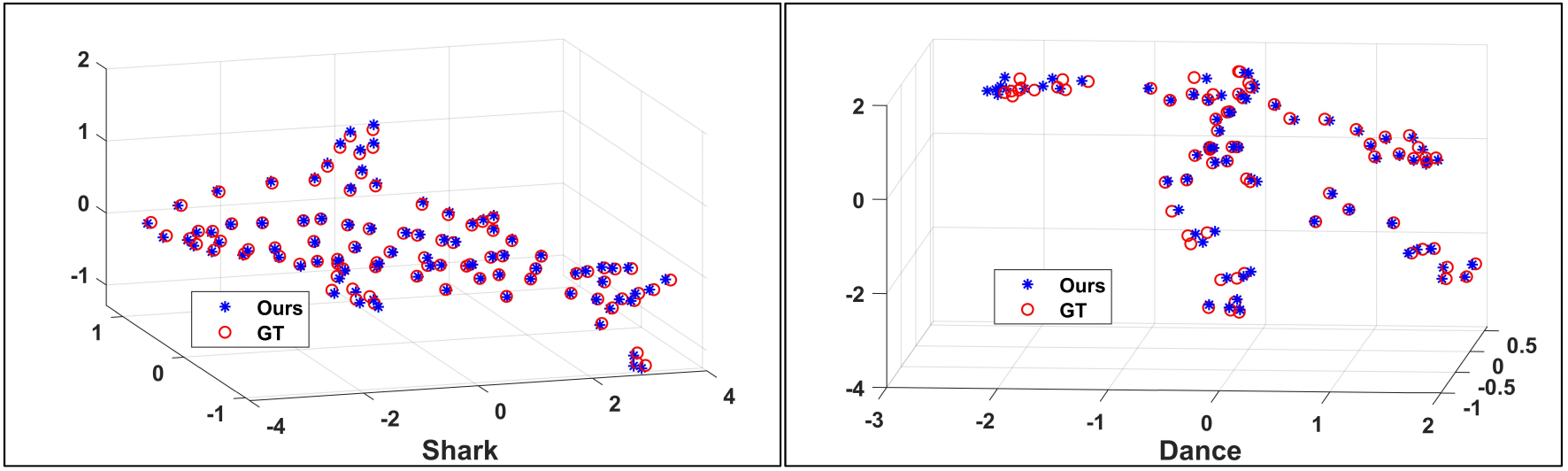}
    \caption{Visualization on MoCap Dataset}
    \label{fig:mocap_vis}
  \end{subfigure}
  \hfill
  \begin{subfigure}{0.98\linewidth}
    \includegraphics[width=1.0\linewidth]{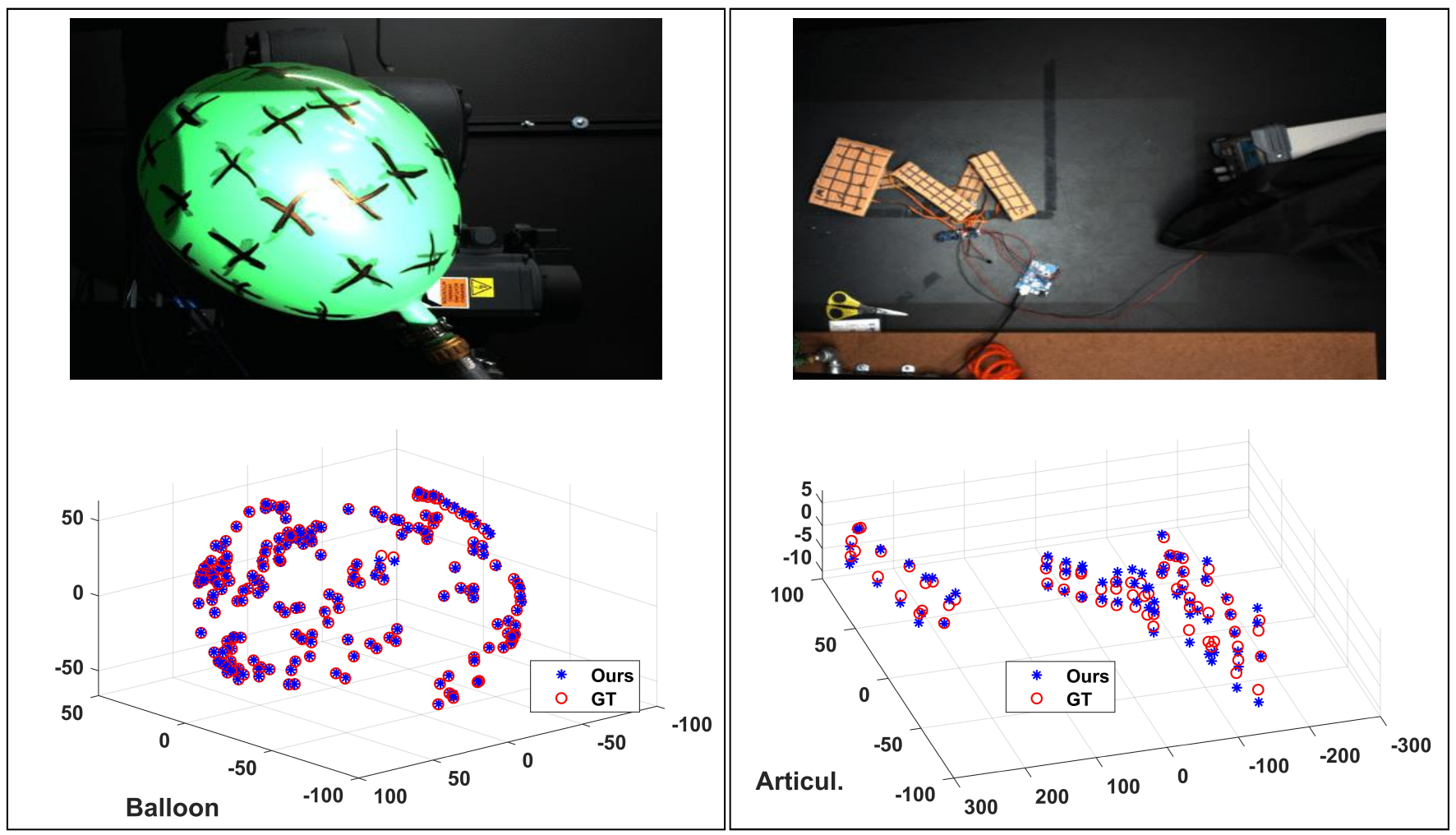}
    \caption{Visualization on NRSfM Challenge Dataset}
    \label{fig:challenge_vis}
  \end{subfigure}
  \hfill
  \begin{subfigure}{0.98\linewidth}
    \includegraphics[width=1.0\linewidth]{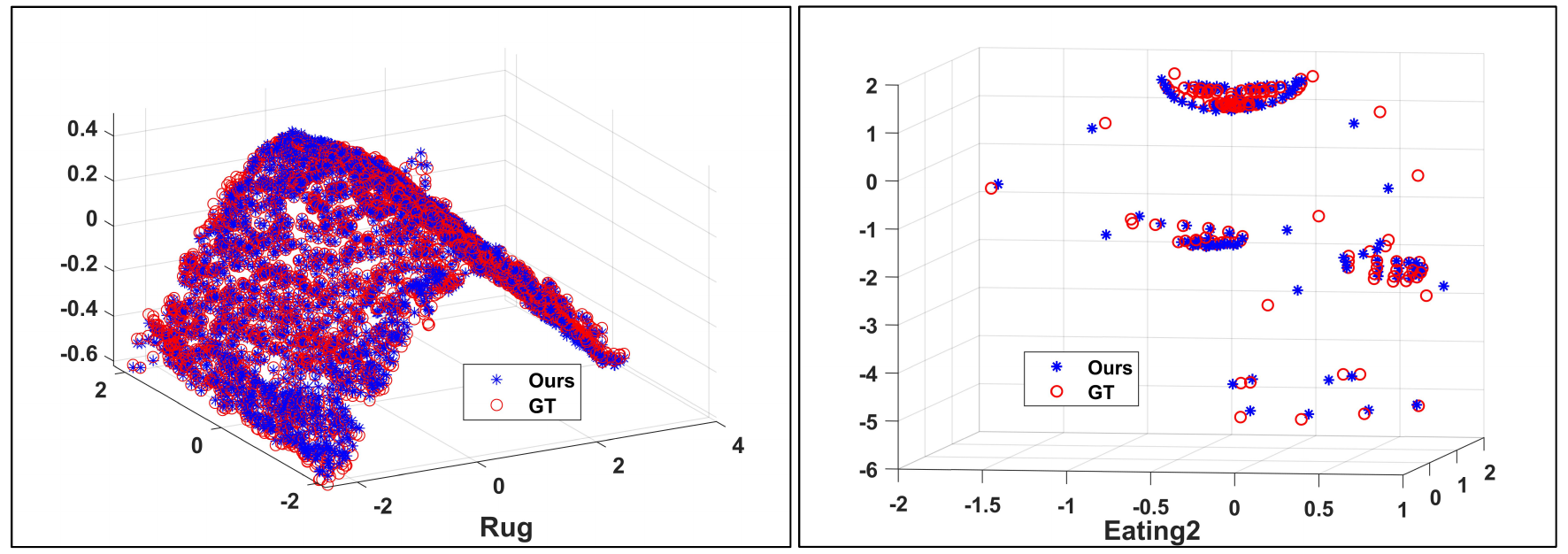}
    \caption{Visualization on Semi-dense and H3WB Dataset}
    \label{fig:dense_vis}
  \end{subfigure}
  \caption{(a) Visualization of Shark and Dance results. (b) Visualization of Balloon and Articulated results. Top row is the image in dataset, and bottom row is the 3D reconstruction shape. (c) Visualization of Rug and Eating2 (Fixed-type) results.}
  \vspace{-1.0em} 
  \label{fig:visualization}
\end{figure}

\textbf{MoCap Benchmark Dataset.} This dataset is a standard benchmark for NRSfM consisting of 8 real sequences. Akhter \etal~\cite{akhter2008nonrigid} introduced five sequences: Drink, Pickup, Yoga, Stretch, and Dance. And the other three, Face, Walking, and Shark, were presented by Torresani \etal~\cite{torresani2008nonrigid}. 
\cref{table_mocap} and~\cref{fig:mocap_vis} demonstrate the reconstruction errors $e_{3d}$ of our method compared to other methods and some visual results, respectively. As shown in~\cref{table_mocap}, our method performs best or second-best across multiple sequences, indicating that our method is able to accommodate diverse types of deformation. Our method also achieves comparable results in sequences such as Shark and Walking, outperforming the pure low-rank constraint methods~\cite{dai2014simple, kumar2020non, kumar2022organic}.

\noindent \textbf{NRSfM Challenge Dataset.}
Jensen \etal~\cite{jensen2021benchmark} recently proposed a new challenging benchmark. This dataset contains five types of non-rigid deformation: Articulated, Balloon, Paper, Stretch, and Tearing. Each subject contains six observation sequences captured by different types of camera motion, \ie, circle, flyby, line, semi-circle, tricky, and zigzag. 
For each subject, we calculate the reconstruction errors under the six camera motions and take the average as the final error for that subject. 
The quantitative comparison with other methods and the qualitative visualization are shown in~\cref{table_challenge} and~\cref{fig:challenge_vis}. Our method achieves the best results on Articul., Stretch, and Tearing, and also has high reconstruction accuracy on Balloon and Paper. 
The numerical results further show the advantages of our method on multi-category deformation reconstruction. 

\begin{table}[t!]
\centering
\caption{Mean normalized 3D reconstruction errors on Semi-dense dataset. ' - ' indicates the estimation failed due to excessive computational overhead.}
\vspace{-0.2cm}
\label{semi-dense}
 \resizebox{1.0\linewidth}{!}{
    \begin{tabular}{c|c|c|c|c|c|c}
    \hline
    Data     & CSF2~\cite{gotardo2011non}      & BMM~\cite{dai2014simple} & CNS~\cite{lee2016consensus}  & R-BMM~\cite{kumar2020non} & Ours & Ours(I)           \\ \hline
    Kinect & 0.0232 & 0.1212 & 0.0453 & 0.0199  & 0.0356    & \textbf{0.0161} \\ 
    Rug & 0.0189 & 0.0109 & - & 0.0135  & \textbf{0.0088}    & \textbf{0.0088} \\ 
    Mat & 0.0620 & 0.1088 & 0.0197 & 0.0285  & 0.0238    & \textbf{0.0182} \\ \hline
    \end{tabular}
}
\end{table}

\begin{table}[t]
\centering
\caption{Comparison of mean normalized 3D reconstruction error with well-known sparse NRSfM methods on H3WB dataset.}
\vspace{-0.2cm}
\label{table_h3wb}
\resizebox{1.0\linewidth}{!}{
\begin{tabular}{c|c|c|c|c|c|c}
    \hline
    \multicolumn{7}{c}{\textbf{Fixed-type camera motion}} \\ \hline
    Data    & {\small CSF2~\cite{gotardo2011non}} & {\small BMM~\cite{dai2014simple}} & {\small PND~\cite{lee2013procrustean}} & {\small CNS~\cite{lee2016consensus}} & {\small R-BMM~\cite{kumar2020non}} & Ours  \\ \hline
    Eating2 & 0.2662 & 0.2084 & 0.2309 & 0.2250 & 0.1779 & \textbf{0.1767} \\ %
    Smoking1  & 0.3534 & 0.2287  & 0.3397  & 0.2982  & 0.2290  & \textbf{0.2040}  \\ %
    Directions  & 0.2953  & 0.2872 & 0.3002 & \textbf{0.2674}  & 0.2964  & \underline{0.2688}  \\ %
    Smoking  & 0.1980 & 0.1759  & 0.4349  & 0.1968  & 0.1760  & \textbf{0.1693}  \\ %
    Waiting2  & 0.2125 & 0.1712 & 0.3306 & 0.1848 & 0.1478 & \textbf{0.1097} \\ \hline
    Mean  & 0.2651 & 0.2143 & 0.3272 & 0.2344 & 0.2054 & \textbf{0.1857} \\ \hline
    \end{tabular}
}
\vspace{-1.0em}
\end{table}

\begin{table}[t]
\centering
\label{table_h3wb_o}
\resizebox{1.0\linewidth}{!}{
\begin{tabular}{c|c|c|c|c|c|c}
    \hline
    \multicolumn{7}{c}{\textbf{One-circle-type camera motion}} \\ \hline
    Data    & {\small CSF2~\cite{gotardo2011non}} & {\small BMM~\cite{dai2014simple}} & {\small PND~\cite{lee2013procrustean}} & {\small CNS~\cite{lee2016consensus}} & {\small R-BMM~\cite{kumar2020non}} & Ours  \\ \hline
    Eating2 & 0.0923 & 0.0665 & 0.0701 & 0.0823 & 0.0633 & \textbf{0.0520} \\ 
    Smoking1  & 0.1230 & 0.0844  & 0.1049  & 0.1073  & 0.0746  & \textbf{0.0738}  \\ 
    Directions  & 0.0787  & 0.0648 & 0.0536 & {0.0621}  & 0.0595  & \textbf{0.0529}  \\ 
    Smoking  & 0.0702 & 0.0492  & 0.0622  & 0.0578  & 0.0363  & \textbf{0.0352}  \\ 
    Waiting2  & \textbf{0.1029} & 0.1081 & 0.1114 & 0.1142 & 0.1117 & \underline{0.1035} \\ \hline
    Mean  & 0.0934 & 0.0746 & 0.0804 & 0.0847 & 0.0691 & \textbf{0.0635} \\ \hline
    \end{tabular}
}
\vspace{-1.0em}
\end{table}

\begin{table}[t]
\centering
\label{table_h3wb_c}
\resizebox{1.0\linewidth}{!}{
\begin{tabular}{c|c|c|c|c|c|c}
    \hline
    \multicolumn{7}{c}{\textbf{Multi-circle-type camera motion}} \\ \hline
    Data    & {\small CSF2~\cite{gotardo2011non}} & {\small BMM~\cite{dai2014simple}} & {\small PND~\cite{lee2013procrustean}} & {\small CNS~\cite{lee2016consensus}} & {\small R-BMM~\cite{kumar2020non}} & Ours  \\ \hline
    Eating2 & 0.0658 & 0.0384 & 0.0395 & 0.0605 & 0.0335 & \textbf{0.0276} \\ 
    Smoking1  & 0.0507 & 0.0486  & 0.0358  & 0.0662  & 0.0332  & \textbf{0.0294}  \\ 
    Directions  & 0.0456  & 0.0569 & 0.0322 & {0.0535}  & 0.0518  & \textbf{0.0315}  \\ 
    Smoking  & 0.0708 & 0.0308  & 0.0292  & 0.0346  & 0.0328  & \textbf{0.0282}  \\ 
    Waiting2  & 0.0892 & 0.0742 & 0.0450 & 0.0873 & 0.0637 & \textbf{0.0390} \\ \hline
    Mean  & 0.0644 & 0.0498 & 0.0363 & 0.0604 & 0.0430 & \textbf{0.0312} \\ \hline
    \end{tabular}
}
\vspace{-1.0em}
\end{table}

\begin{table*}[t]
\centering
\scriptsize
\caption{\centering Comparison of reconstruction errors with and without SWNN module. The table shows the comparison results on the MoCap dataset and the Articul.(circle), Articul.(flyby), Stretch(zigzag) sequences in NRSfM Challenge dataset.}
\label{swnn_exp}
\vspace{-0.2cm}
\resizebox{0.9\linewidth}{!}{
\begin{tabular}{c|c|c|c|c|c|c|c|c|c}
\hline
Method    & Drink          & Pickup         & Yoga           & Stretch        & Dance          & Face                    & Articul.(c)     & Articul.(f)  & Stretch(z)    \\ \hline
Ours(w/)  & \textbf{0.0031} & \textbf{0.0126} & \textbf{0.0109} & \textbf{0.0114} & \textbf{0.0921} & \textbf{0.0144}  & \textbf{2.3442} & \textbf{3.8229} & \textbf{0.6460} \\ %
Ours(w/o) & 0.0050          & 0.0137          & 0.0120          & 0.0132          & 0.0932          & 0.0161                   & 3.5626          & 4.8042     & 0.7341    \\ \hline
\end{tabular}}
\vspace{-1.0em}
\end{table*}

\noindent \textbf{Semi-dense Dataset.} We evaluated our method on the kinect paper, rug, table mat datasets~\cite{parashar2017isometric}($191, 159, 60$ frames and $1503, 3912, 1500$ keypoints respectively). We compared our method with other well-known sparse NRSfM methods (PND~\cite{lee2013procrustean} fails on this dataset due to excessive computational overhead caused by the large number of points), see the results in~\cref{semi-dense} and~\cref{fig:dense_vis}. We denote our method that initializes $\mathbf{R}_{p}$ using the modified algorithm in~\cite{kumar2020non} as Our(I). Our proposed model is also effective in reducing the reconstruction error on semi-dense dataset. Our method can significantly reduce the effect of poor camera motion on shape estimation and also further improve reconstruction accuracy with better motion initialization.

\noindent \textbf{H3WB Dataset.} Human3.6M 3D WholeBody (H3WB) dataset~\cite{zhu2023h3wb} is an entire human body 3D dataset, including face, hands, body, and feet. H3WB is an extension of Human3.6M~\cite{ionescu2013human3} and contains multiple categories of common life actions such as Directions, Eating, and Smoking. The human body in H3WB is annotated with $133$ keypoints, $17$ for body, $6$ for feet, $68$ for face, and $42$ for hands. We selected five sequences in H3WB and processed them to obtain a new evaluation dataset: Eating2, Smoking1, Directions, Smoking, Waiting2 ($185, 265, 245, 180$ and $335$ frames, respectively). We simulated three types of camera motion: \textbf{fixed-type}, \textbf{one-circle-type}, and \textbf{multi-circle-type}, then generated three sets of 2D observations for testing by orthogonal projection, see {Supp.~Sec~4} for more details\footnote{The results we reported in \red{conference version} of the paper is actually just the reconstruction error on \red{Fixed-type camera motion} setting, we added more results here to domenstrate the effectiveness of our method.}. As manifested in~\cref{table_h3wb}, our method shows better reconstruction results than other prior-free and procrustean-based methods (Visualization illustrated in~\cref{fig:dense_vis}).
When the camera captures the object from as many different viewpoints as possible (multi-circle-type), our method is able to reconstruct the 3D structures very accurately. Moreover, when the camera is fixed (fixed-type), ours can also give more reliable results.

\begin{figure}
  \centering
  \includegraphics[width=1.0\linewidth]{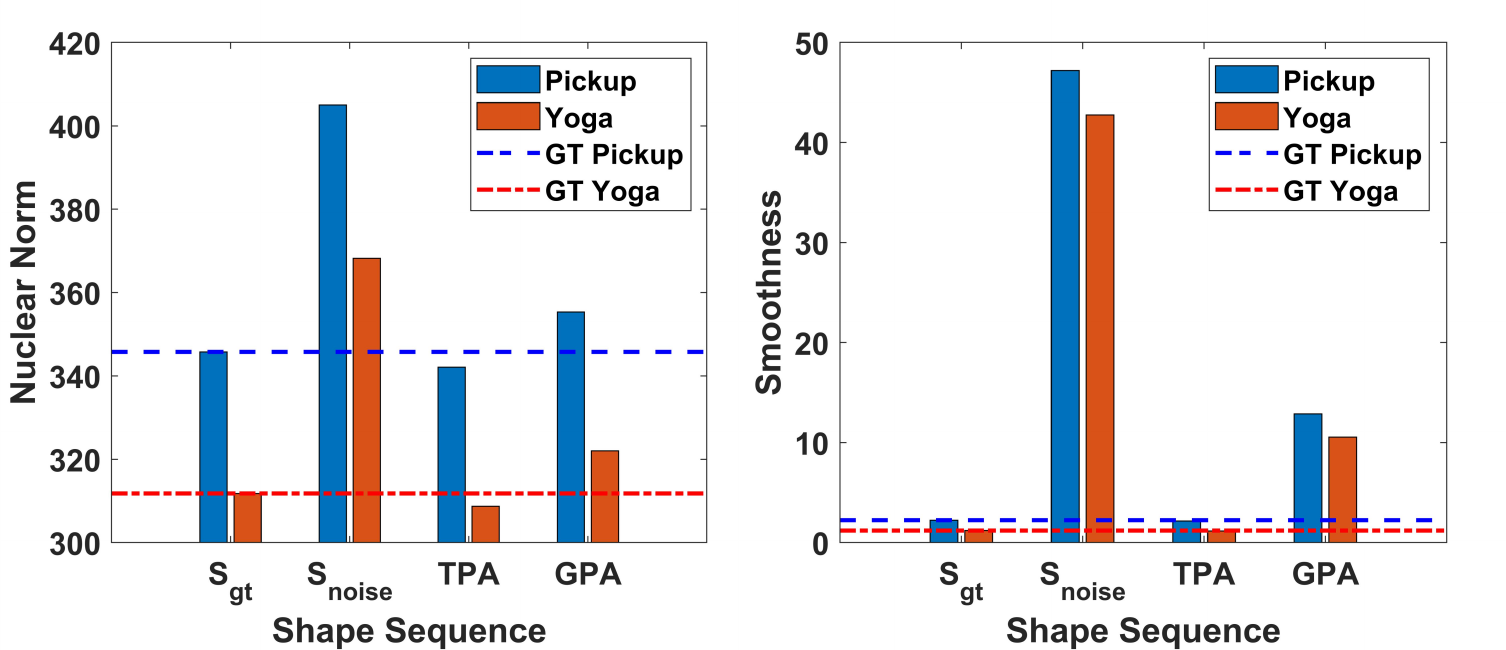}
  \caption{Shape alignment test for the TPA module. We compare the low-rank and smoothing properties of GT non-rigid sequences, randomly rotated disrupted sequences, and sequences aligned by TPA or GPA. The results show that the TPA-aligned sequences have more similar properties to the GT sequences.}\vspace{-1.0em}
  \label{fig:spa_exp}
\end{figure}

\subsection{Ablation Study}\label{ablation_sec}
\vspace{-0.2em}

\noindent \textbf{Role of TPA Module.} First, we verify the alignment function of TPA module \eqref{TPA_module} on Pickup and Yoga sequences. We randomly sample Lie algebra $p_{n}\in \mathbb{R}^{3\times F}$ from Gaussian noise $\mathcal{N}(0, 0.1)$, and then map it to a 3D rotation matrix $\mathbf{R}_{d}\in \mathbb{R}^{3F \times 3F}$ using Rodrigues' rotation formula. We define the shape sequence disrupted by random rotations as $\mathbf{S}_{noise}=\mathbf{R}_{d}\mathbf{S}_{gt}$, where $\mathbf{S}_{gt} \in \mathbb{R}^{3F \times P}$ is the GT sequence in the world coordinate. We realigned the sequence with the TPA module and compared the result with GPA~\cite{lee2013procrustean}. \cref{fig:spa_exp} compares the nuclear norm $(\ie, \| g(\cdot) \|_{\ast})$ and the first-order smoothness~\cite{dai2014simple}  of $\mathbf{S}_{gt}, \mathbf{S}_{noise}$ and the sequences aligned by the two algorithms, showing that the TPA-aligned sequence has the most similar properties to the GT sequence.

We also test the stability of the TPA module for camera motion initialization. We used the same approach to sample perturbed rotations $\mathbf{R}_{d}$ with standard deviation $\sigma$ of 0.1, 0.2, and 0.5. Adding ambiguity $\mathbf{R}_{d}$ to the camera motion initialized by BMM~\cite{dai2014simple}, we then compare the four methods BMM, R-BMM, BMM+Smooth (Add first-order smoothness constraint), BMM+TPA. For each $\sigma$, we experimented five times and calculated the mean, the results are shown in~\cref{spa_exp2}. 
Compared to other methods, the TPA module maintains its effectiveness even when a large deviation has been applied to the camera motion.
This shows that our method can effectively deal with the rotation ambiguity in camera motion estimation and can improve the effectiveness and stability of the smoothing constraint.

\begin{table}[t]
\centering
\caption{Experiment for the effectiveness of the TPA module in correcting camera motion. Adjustment for weight $\mu_{3}$ in brackets.}
\vspace{-0.2cm}
\label{spa_exp2}
 \resizebox{1.0\linewidth}{!}{
    \begin{tabular}{c|c|c|c|c}
    \hline
    SD           & BMM~\cite{dai2014simple}  & R-BMM~\cite{kumar2020non} & BMM+Smooth & BMM+TPA           \\ \hline
    $\sigma=0.1$ & 0.0493 & 0.0507 & 0.0845     & \textbf{0.0237}($1e^{-1}$) \\ 
    $\sigma=0.2$ & 0.0845 & 0.0769 & 0.1149     & \textbf{0.0592}($1e^{-2}$) \\ 
    $\sigma=0.5$ & 0.1743 & 0.1681 & 0.1905     & \textbf{0.1665}($1e^{-3}$) \\ \hline
    \end{tabular}
}
\vspace{-1.0em}
\end{table}

\begin{figure}
  \centering
  \begin{subfigure}{0.49\linewidth}
    \includegraphics[width=1.0\linewidth]{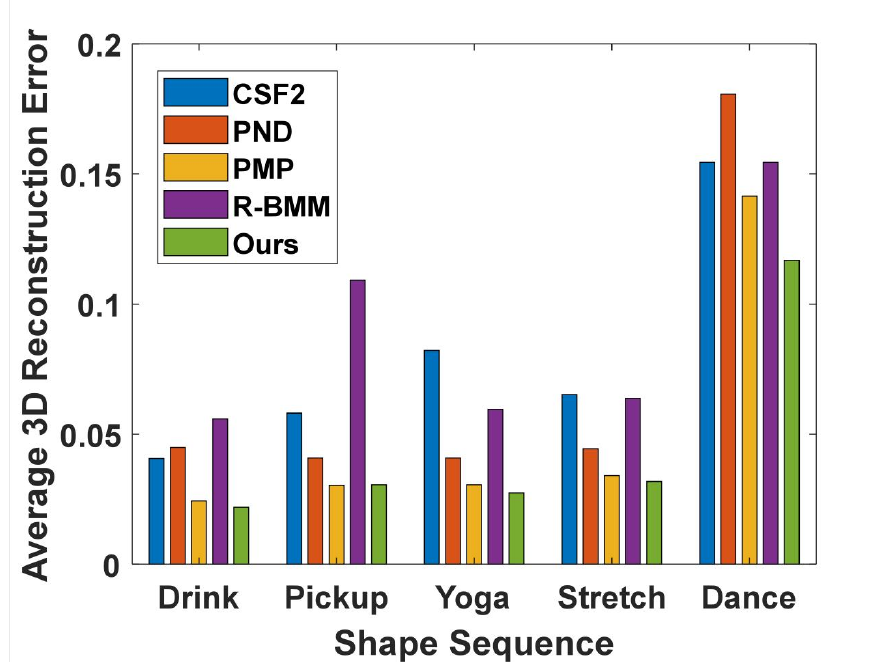}
    \caption{$e_{3d}$ with noise}
    \label{fig:noise}
  \end{subfigure}
  \hfill
  \begin{subfigure}{0.49\linewidth}
    \includegraphics[width=1.0\linewidth]{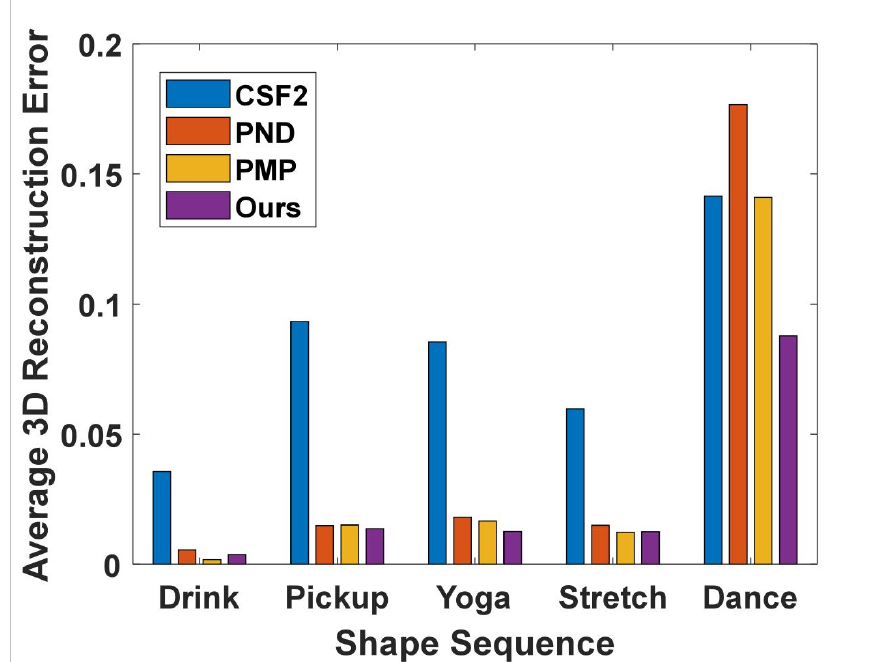}
    \caption{$e_{3d}$ with occlusion}
    \label{fig:occl}
  \end{subfigure}
  \caption{(a) Shape errors on noisy sequences. (b) Experiment for 3D reconstruction on missing data.}\vspace{-1.0em}
  \label{fig:abla_occl_noise}
\end{figure}

\noindent\textbf{Effectiveness of SWNN.} We compared the two cases of using and not using (with $\mathbf{\Lambda} = \mathbf{I}$) the SWNN module in model \eqref{whole_model1}. The comparison results are reported in~\cref{swnn_exp}. SWNN improves the accuracy of 3D shape recovery in most deformation settings. 
The accuracy improvement mainly depends on the segmentation result and the smoothing property of the sequence, for more analysis see Supp.~Sec~3.

\noindent\textbf{Performance on Noisy and Missing Data.} We follow the approaches in~\cite{lee2013procrustean} to add noise and occlusion to the data. For missing data, we first complete the measurement matrix by~\cite{cabral2013unifying} and then run our method. \cref{fig:abla_occl_noise} shows that our method performs better than others in most cases\footnote{See Supp. for more tests on TPA, SWNN and Missing Data.}.

\vspace{-0.3em}
\section{Conclusion}
\vspace{-0.2em}
In this paper, we have proposed a spatial-temporal modeling framework for NRSfM. 
We proposed a Temporally-smooth Procrustean Alignment module to tackle the rotation ambiguity. 
Furthermore, we introduced a spatial-weighted nuclear norm model with the shape proxy strategy to ensure the effectiveness of low-rank constraint in coping with severe spatially-variant deformations.
Our method achieves superior 3D reconstruction performance across a wide range of deformation sequences. 
Moreover, there is still much research to do in the future on more accurate trajectory segmentation and explicit modeling of complex deformation.

\noindent\textbf{Acknowledgements: }This research was supported in part by the National Natural Science Foundation of China (62271410), and the Fundamental Research Funds for the Central Universities.

{
    \small
    \bibliographystyle{ieeenat_fullname}
    \bibliography{main}
}
\setcounter{section}{0}

\clearpage
\setcounter{page}{1}
\maketitlesupplementary

\begin{abstract}
In this supplementary material, we provide additional materials in the following aspects: 1) solution of the Temporally-smooth Procrustean Alignment (TPA) model and verification of its validity; 2) detailed solution of the ADMM model; 3) additional explanations for spatially-variant deformation modeling; and 4) additional details of the experimental setup, supplementation of missing data experiments, and more visualization results.
\end{abstract}

\vspace{-1.0em}
\section{Temporally-smooth Procrustean Alignment}
In this section, we first provide details in solving our proposed {Temporally-smooth Procrustean Alignment (TPA)} module and then report additional experimental results.

\subsection{Algorithm for TPA}\label{algo_for_spa}
Denote the 3D shape sequence to be aligned as $\{\mathbf{S}_{i}\}_{i=1}^{F}$, we set the optimization target of TPA as:
\begin{equation}\label{smooth_align_target}
    \mathcal{L}^{Q}_{tpa} = \frac{1}{2}\sum_{i=1}^{F-1}{\left \| \mathbf{Q}_{i}\mathbf{S}_{i} - \mathbf{Q}_{i+1}\mathbf{S}_{i+1} \right \|}^{2}_{F},
\end{equation}
where we initialize $\mathbf{Q}_{i}=\mathbf{I}_{3}$, and $\mathbf{I}_{3} \in \mathbb{R}^{3\times 3}$ denotes the identity matrix. We solve this optimization problem in a frame-by-frame manner. When $1<i<F$, we transfer the optimization target as:
\begin{align}
    \nonumber \mathcal{L}^{Q_{i}}_{tpa} & = \frac{1}{2}\sum_{k=i-1}^{i}\sum_{j=1}^{P}{\left \| \mathbf{Q}_{k}\mathbf{s}_{k,j} - \mathbf{Q}_{k+1}\mathbf{s}_{k+1,j} \right \|}^{2}_{2} \\ 
    \nonumber & = \frac{1}{2}\sum_{j=1}^{P}({ \left \| \mathbf{Q}_{i-1}\mathbf{s}_{i-1,j} - \mathbf{Q}_{i}\mathbf{s}_{i,j} \right \| }^{2}_{2} \\ 
    &\ \ \ \ \ \ \ \ \ \ \  + \left \| \mathbf{Q}_{i}\mathbf{s}_{i,j} - \mathbf{Q}_{i+1}\mathbf{s}_{i+1,j} \right \|^{2}_{2}),
\end{align}
where $\mathbf{s}_{i,j}$ is the $j$-th column of 3D shape $\mathbf{S}_{i}$. 
Taking the Lie algebra $\phi_{i}$ corresponding to the rotation matrix $\mathbf{Q}_{i}$ as the optimization variable, the gradient of model \eqref{smooth_align_target} can be calculated as:
\begin{equation}\label{eq:grad_for_spa}
    \mathbf{g}_{i} = \frac{\partial \mathcal{L}^{Q_{i}}_{tpa}}{\partial \phi_{i}} = \frac{\partial \frac{1}{2}\sum_{j=1}^{P}{(\left \| \mathbf{r}_{i,j}^{(1)} \right \|^{2}_{2}+\left \| \mathbf{r}_{i,j}^{(0)} \right \|^{2}_{2})}}{\partial \phi_{i}}
\end{equation}
\begin{align}
   \nonumber & = \sum_{j=1}^{P}{\left[(\frac{\partial \mathbf{r}_{i,j}^{(1)}}{\partial \phi_{i}})^T\mathbf{r}_{i,j}^{(1)} + ( \frac{\partial \mathbf{r}_{i,j}^{(0)}}{\partial \phi_{i}})^T\mathbf{r}_{i,j}^{(0)} \right]} \\
    \nonumber & = \sum_{j=1}^{P}{\left [ \left ( \mathbf{Q}_{i}\mathbf{s}_{i,j} \right )^{\wedge } \right ]^{T}}\left ( r_{i,j}^{(1)} - r_{i,j}^{(0)} \right ),
\end{align}
where $\mathbf{r}_{i,j}^{(m)}\!=\!\mathbf{Q}_{i-m}\mathbf{s}_{i-m,j}\!-\!\mathbf{Q}_{i-m+1}\mathbf{s}_{i-m+1,j}$ is the residual vector, and $\left ( \mathbf{Q}_{i}\mathbf{s}_{i,j} \right )^{\wedge }$ is an approximation to the derivative of the residual $\mathbf{r}_{i,j}^{(m)}$ with respect to the Lie algebra $\phi_{i}$. $\left ( \mathbf{a} \right )^{\wedge}$ is the skew-symmetric matrix of vector $\mathbf{a}$, \ie:
\begin{equation}
    \left ( \begin{bmatrix}
 a_{1}\\
 a_{2}\\
a_{3}
\end{bmatrix} \right )^{\wedge} =\begin{bmatrix}
 0&  -a_{3}&a_{2} \\
 a_{3}&  0& -a_{1}\\
 -a_{2}&  a_{1}&0
\end{bmatrix}.
\end{equation}
Similarly, when $i = 1~or~F$, the gradient $\mathbf{g}_{i}$ can be calculated as:
\begin{align}\label{eq:grad_for_spa_F}
   \mathbf{g}_{i} =\left\{\begin{matrix}
  -\sum_{j=1}^{P}{\left [ \left ( \mathbf{Q}_{i}\mathbf{s}_{i,j} \right )^{\wedge } \right ]^{T}}r_{i,j}^{(0)},~~~i=1,\\
\sum_{j=1}^{P}{\left [ \left ( \mathbf{Q}_{i}\mathbf{s}_{i,j} \right )^{\wedge } \right ]^{T}}r_{i,j}^{(1)},~~~~~~~i=F.
\end{matrix}\right. 
\end{align}
The gradient $\mathbf{g}_{i}$ can be expressed as the product of the Jacobian matrix and the residual, \ie, $\mathbf{g}_{i}  = \mathbf{J}_{i}^{T}\mathbf{r}_{i}$. $\mathbf{J}_{i} \in \mathbb{R}^{3P \times 3}$ is the Jacobian matrix as follows:
\begin{equation}\label{Jacobi}
    \mathbf{J}_{i} = \left [ \left (\mathbf{Q}_{i}\mathbf{s}_{i,1}  \right )^{\wedge } \cdots \left (\mathbf{Q}_{i}\mathbf{s}_{i,P}  \right )^{\wedge } \right ]^{T},
\end{equation}
where $\mathbf{r}_{i}$ is the column vector stacked from residuals $\{ r_{i,j}^{(1)} - r_{i,j}^{(0)} \}_{j=1}^{P}$ or $\{r_{i,j}^{(m)} \}_{j=1}^{P}$. After obtaining the Jacobian matrix, we can compute the approximation of the second-order Hessian matrix as $\mathcal{H}_{i} = \mathbf{J}_{i}^{T}\mathbf{J}_{i}$ and use numerical optimization methods such as the Levenberg-Marquardt (LM) algorithm~\cite{nocedal2006numerical} to solve for the rotation matrix $\mathbf{Q}=diag(\mathbf{Q}_{i})$.

\begin{figure*}
  \centering
  \includegraphics[height=21cm,width=0.95\linewidth]{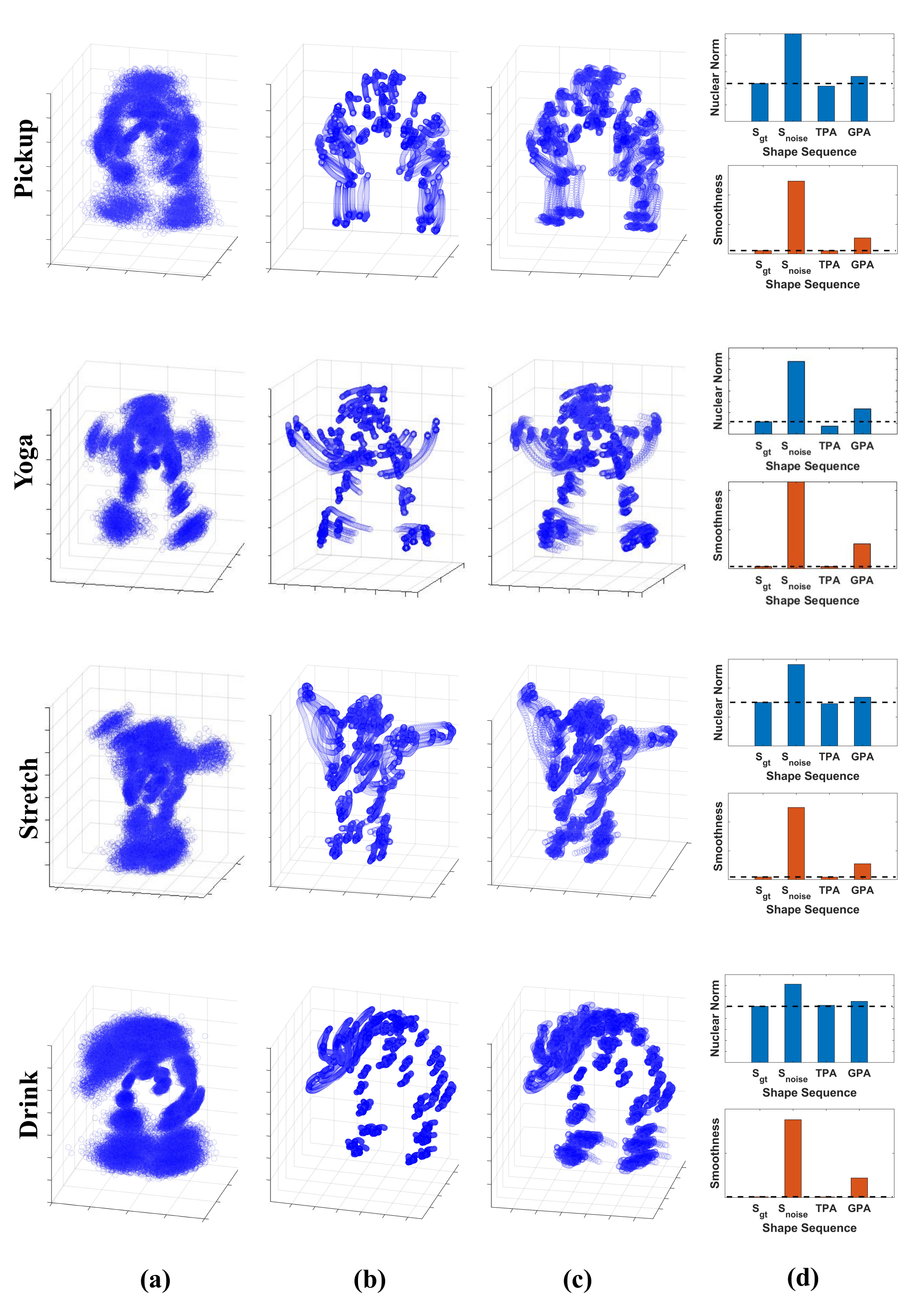}
  \caption{Experiments for testing the alignment capability of TPA module. (a) Visualization of the shape sequences $\mathbf{S}_{noise}$. (b) Visualization of the TPA-aligned shape sequences. (c) Visualization of the GPA-aligned shape sequences. (d) The low-rank and smoothing properties of GT non-rigid sequences $\mathbf{S}_{gt}$, randomly rotated disrupted sequences $\mathbf{S}_{noise}$, and sequences aligned by TPA or GPA.}
  \label{fig:supp_tpa}
\end{figure*}

\subsection{Experiments for TPA}\label{tpa_exp_setting}
In this subsection, we verify the effectiveness of the TPA module. Let's revisit the setting of the experiments. We randomly sample Lie algebra $p_{n}\in \mathbb{R}^{3\times F}$ from Gaussian distribution $\mathcal{N}(0, 0.1)$, and then map it to a block diagonal matrix $\mathbf{R}_{d}\in \mathbb{R}^{3F \times 3F}$ consisting of 3D rotations using the Rodrigues' rotation formula. We define the shape sequence disrupted by random rotations as $\mathbf{S}_{noise}=\mathbf{R}_{d}\mathbf{S}_{gt}$, where $\mathbf{S}_{gt} \in \mathbb{R}^{3F \times P}$ is the GT sequence in the world coordinate. We realign the sequence using the TPA and GPA~\cite{lee2013procrustean} modules and compare their results.
We performed these experiments on the Pickup, Yoga, Stretch, and Drink sequences in MoCap benchmark and the results are shown in~\cref{fig:supp_tpa}.

We centralized all the 3D shapes in the sequence and plotted them in the same coordinate. Column (a) in~\cref{fig:supp_tpa} is the disrupted sequence $\mathbf{S}_{noise}$. Columns (b) and (c) are sequences after TPA or GPA alignment, respectively. Column (d) shows the low-rank and smoothing properties of real sequences $\mathbf{S}_{gt}$, noisy sequences $\mathbf{S}_{noise}$, TPA-aligned sequences, and GPA-aligned sequences. \cref{fig:supp_tpa} demonstrates that the TPA module can efficiently align the 3D shape sequences. Compared with GPA, the TPA-aligned sequences are more similar to the real 3D shape sequences in terms of their low-rank and smoothing properties.

Non-rigid deformation mixed with rigid motion cannot be accurately estimated using the low-rank constraint. As manifested in~\cref{fig:supp_tpa} (d), shape sequences after alignment by the TPA module have lower nuclear norm and smoothness. From one hand, this demonstrates that the TPA module can effectively separate rigid motion from non-rigid deformation. From the other hand, the properties of the TPA-aligned sequence make it can be better recovered by the low-rank and smoothing constraints. In summary, we utilize the low-rank constraint to recover the TPA-aligned shape sequence, rather than directly applying the low-rank constraint under the coordinate defined by the estimated camera motion~\cite{dai2014simple, kumar2020non, kumar2022organic}.

\section{ADMM Model Solution}
\subsection{Model without Occlusion}
In this section, we provide the completed solution of our proposed method. Let's first recall the Lagrangian model to be solved:
\vspace{-0.7em}
\begin{equation}
 \resizebox{0.45\textwidth}{!}{$
 \begin{aligned}\label{lagrangian_formula}
     \min_{\Omega} ~ \mathcal{L} & =  \frac{\mu_{1}}{2}\left \| \mathbf{W}\!-\!\mathbf{\Pi}\mathbf{S} \right \|^{2}_{F}  + \mu_{2}\left \| \mathbf{\breve{S}}^{\sharp} \right \|_{*} + \frac{\mu_{3}}{2}\sum_{i=1}^{F-1}\left \| \mathbf{Q}_{i}\mathbf{\tilde{S}}_{i}\!-\!\mathbf{Q}_{i+1}\mathbf{\tilde{S}}_{i+1} \right \|_{F}^{2} \\
     & + \frac{\beta}{2}\left \| \mathbf{\breve{S}}^{\sharp} - g(\mathbf{\hat{S}}\mathbf{\Lambda})  \right \|_{F}^{2}+\left \langle \mathbf{Y}_{1}, \mathbf{\breve{S}}^{\sharp} - g(\mathbf{\hat{S}}\mathbf{\Lambda}) \right \rangle \\
     & + \frac{\beta}{2}\left \| \mathbf{\hat{S}} - \mathbf{Q}\mathbf{\tilde{S}} \right \|_{F}^{2}+\left \langle \mathbf{Y}_{2},\mathbf{\hat{S}} - \mathbf{Q}\mathbf{\tilde{S}} \right \rangle \\
     & + \frac{\beta}{2}\left \| \mathbf{\tilde{S}} - \mathbf{R}_{p}\mathbf{{S}} \right \|_{F}^{2}+\left \langle \mathbf{Y}_{3}, \mathbf{\tilde{S}} - \mathbf{R}_{p}\mathbf{{S}} \right \rangle , 
 \end{aligned}$}
 \raisetag{6\baselineskip}
\end{equation}
where $\Omega =  \{\mathbf{S,\tilde{S},\hat{S},\breve{S}^{\sharp},Q} \}$ denotes the variables to be updated, $\{\mathbf{Y}_{n}\}_{n=1}^{3}$ are the Lagrange multipliers. We then give the update formula for each optimization variable in $\Omega$.

\noindent\textbf{Solution for $\mathbf{\breve{S}}^{\sharp}$.} Selecting all the optimization terms in model \eqref{lagrangian_formula} that are related to $\mathbf{\breve{S}}^{\sharp}$, the optimization model for $\mathbf{\breve{S}}^{\sharp}$ is obtained as:
\begin{equation}
    \begin{aligned}
    \mathbf{\breve{S}}^{\sharp} & = \arg \min_{\mathbf{\breve{S}}^{\sharp}}~\mu_{2}\left \| \mathbf{\breve{S}}^{\sharp} \right \|_{*} + \frac{\beta}{2}\left \| \mathbf{\breve{S}}^{\sharp} - g(\mathbf{\hat{S}}\mathbf{\Lambda}) \right \|_{F}^{2} \\
         & + \left \langle \mathbf{Y}_{1},\mathbf{\breve{S}}^{\sharp} - g(\mathbf{\hat{S}}\mathbf{\Lambda}) \right \rangle \\
         & = \arg \min_{\mathbf{\breve{S}}^{\sharp}}~\frac{\beta}{2}\left \| \mathbf{\breve{S}}^{\sharp} - \left ( g(\mathbf{\hat{S}}\mathbf{\Lambda})- \frac{1}{\beta}\mathbf{Y}_{1} \right ) \right \|_{F}^{2} + \mu_{2}\left \| \mathbf{\breve{S}}^{\sharp} \right \|_{*}.
    \end{aligned}
\end{equation}
We use the method in~\cite{kumar2020non} to obtain the closed-form solution. We first define the soft-thresholding function $\mathcal{S}_{\tau}(\sigma)=sign(\sigma)\max(\left | \sigma \right | - \tau, 0 )$. Then the closed-form solution $\mathbf{\breve{S}}^{\sharp}$ can be given by:
\begin{equation}
    \begin{aligned}\label{solution_of_X}
    & \mathbf{\breve{S}}^{\sharp} = \mathbf{U}\mathcal{S}_{\frac{\Theta\mu_{2}}{\beta}}(\mathbf{\Sigma})\mathbf{V}^{T}, \\
    & \mathbf{U}, \mathbf{\Sigma}, \mathbf{V} = \mathrm{SVD}\left ( g(\mathbf{\hat{S}}\mathbf{\Lambda})- \frac{1}{\beta}\mathbf{Y}_{1} \right ),
    \end{aligned}
\end{equation}
where $\Theta$ is the weight set for different singular values. The larger the singular value tends to be the more significant, and it should correspond to a smaller weight. The weight $\Theta$ in~\cite{kumar2020non} is set as:
\begin{equation}
    \Theta_{j} = \frac{\xi}{\sigma_{j}(\mathbf{X}) + \gamma},
\end{equation}
where $\sigma_{j}(\mathbf{X})$ is the singular value of $\mathbf{X}$, $\xi$ is a positive number and $\gamma=1e^{-6}$. Here we use the relative magnitude of the singular values to adjust the weight setting, \ie:
\begin{equation}
    \tilde{\Theta}_{j} = \left\{\begin{aligned}
 & \frac{\xi \cdot \Theta_{j}}{\sum_{i=1}^{K_{s}}{\Theta_{i}}},~~~1\le j \le K_{s},\\
& ~~~~~~~0,~~~~~~~~~~K_{s}< j.
\end{aligned}\right.
\end{equation}
We set the shape basis dimension $K_{s}$ according to the assumption of linear basis combination and truncate the singular values. We can better measure the importance of different singular values by adjusting the weights through normalization.

\noindent\textbf{Solution for $\mathbf{\hat{S}}$.} The solution model for the optimization variable $\mathbf{\hat{S}}$ can be expressed as follows:
\begin{equation}
    \begin{aligned}
        \mathbf{\hat{S}} & = \arg \min_{\mathbf{\hat{S}}}~\frac{\beta}{2}\left \| \mathbf{\breve{S}}^{\sharp} - g(\mathbf{\hat{S}}\mathbf{\Lambda}) \right \|_{F}^{2} + \left \langle \mathbf{Y}_{1},\mathbf{\breve{S}}^{\sharp} - g(\mathbf{\hat{S}}\mathbf{\Lambda}) \right \rangle \\
        & + \frac{\beta}{2}\left \| \mathbf{\hat{S}} - \mathbf{Q}\mathbf{\tilde{S}} \right \|_{F}^{2}+\left \langle \mathbf{Y}_{2},\mathbf{\hat{S}} - \mathbf{Q}\mathbf{\tilde{S}} \right \rangle,
    \end{aligned}
\end{equation}
where $g(\cdot)$ is an invertible linear operator, so we can rewrite the above equation in a more easily solvable form as:
\begin{equation}
    \begin{aligned}\label{S_hat}
        & \mathbf{\hat{S}} = \arg \min_{\mathbf{\hat{S}}}~\frac{\beta}{2}\left \| \mathbf{\hat{S}} - \mathbf{Q}\mathbf{\tilde{S}} \right \|_{F}^{2}+\left \langle \mathbf{Y}_{2},\mathbf{\hat{S}} - \mathbf{Q}\mathbf{\tilde{S}} \right \rangle \\
        & + \frac{\beta}{2}\left \| g^{-1}(\mathbf{\breve{S}}^{\sharp}) - \mathbf{\hat{S}}\mathbf{\Lambda} \right \|_{F}^{2} + \left \langle g^{-1}(\mathbf{Y}_{1}), g^{-1}(\mathbf{\breve{S}}^{\sharp}) - \mathbf{\hat{S}}\mathbf{\Lambda} \right \rangle.
    \end{aligned}
\end{equation}
The closed-form solution of $\mathbf{\hat{S}}$ can be computed by taking the derivative of the model \eqref{S_hat} and equating it to zero:
\begin{equation}
    \begin{aligned}\label{solution_for_S_hat}
         \mathbf{\hat{S}} ( \mathbf{I}_{P} &+ \mathbf{\Lambda}\mathbf{\Lambda}^{T} ) = \\ & \mathbf{Q\tilde{S}} - \frac{1}{\beta}\mathbf{Y}_{2} + g^{-1}(\mathbf{\breve{S}}^{\sharp})\mathbf{\Lambda}^{T} + \frac{1}{\beta}g^{-1}(\mathbf{Y}_{1})\mathbf{\Lambda}^{T}.
    \end{aligned}    
\end{equation}

\noindent\textbf{Solution for $\mathbf{\tilde{S}}$.} First, we define the first-order smoothing matrix $\mathbf{H} \in \mathbb{R}^{3F\times 3F}$ as in~\cite{dai2014simple}:
\begin{equation}
    \mathbf{H}_{i,j} = \left\{\begin{aligned}
&1, ~~~~~~~j = i,i = 1,..., 3(F-1),\\ 
&-1, ~~j = i + 3,i = 1,..., 3(F-1),\\ 
&0, ~~~~~~\mathrm{Otherwise}.
\end{aligned}\right.
\end{equation}
By introducing the matrix $\mathbf{H}$, we can denote the TPA module equivalently as:
\begin{equation}
    \sum_{i=1}^{F-1}\left \| \mathbf{Q}_{i}\mathbf{\tilde{S}}_{i} -\mathbf{Q}_{i+1}\mathbf{\tilde{S}}_{i+1} \right \|_{F}^{2} = \left \| \mathbf{H}\mathbf{Q}\mathbf{\tilde{S}} \right \|_{F}^{2}.
\end{equation}
Then the optimization model for the variable $\mathbf{\tilde{S}}$ can be expressed as follows:
\begin{equation}
    \begin{aligned}\label{S_tilde}
        \mathbf{\tilde{S}} & = \arg \min_{\mathbf{\tilde{S}}}~\frac{\mu_{3}}{2}\left \| \mathbf{H}\mathbf{Q}\mathbf{\tilde{S}} \right \|_{F}^{2} \\
        & + \frac{\beta}{2}\left \| \mathbf{\hat{S}} - \mathbf{Q}\mathbf{\tilde{S}} \right \|_{F}^{2}+\left \langle \mathbf{Y}_{2},\mathbf{\hat{S}} - \mathbf{Q}\mathbf{\tilde{S}} \right \rangle \\
        & + \frac{\beta}{2}\left \| \mathbf{\tilde{S}} - \mathbf{R}_{p}\mathbf{{S}} \right \|_{F}^{2}+\left \langle \mathbf{Y}_{3}, \mathbf{\tilde{S}} - \mathbf{R}_{p}\mathbf{{S}} \right \rangle.
    \end{aligned}
\end{equation}
Calculating the derivative of model \eqref{S_tilde} and equating it to zero yields the closed-form solution of $\mathbf{\tilde{S}}$ as follows:
\begin{equation}
    \begin{aligned}\label{solution_for_S_tilde}
        (\frac{\mu_{3}}{\beta}\mathbf{Q}^{T}\mathbf{H}^{T}\mathbf{H}\mathbf{Q} & + 2\mathbf{I}_{3F})\mathbf{\tilde{S}} = \\
        & \mathbf{Q}^{T}\mathbf{\hat{S}} + \frac{1}{\beta}\mathbf{Q}^{T}\mathbf{Y}_{2} + \mathbf{R}_{p}\mathbf{{S}} - \frac{1}{\beta}\mathbf{Y}_{3}.
    \end{aligned}    
\end{equation}

\begin{algorithm}[t]
	\renewcommand{\algorithmicrequire}{\textbf{Input:}}
	\renewcommand{\algorithmicensure}{\textbf{Output:}}
	\caption{ADMM Optimization Algorithm}
	\label{alg2}
	\begin{algorithmic}[1]
		\STATE \textbf{Input:} Initialize $\mathbf{S},\mathbf{\tilde{S}},\mathbf{\hat{S}},\mathbf{\breve{S}^{\sharp}}, \mathbf{R}_{p}, \mathbf{Q}, \mathbf{\Lambda}, \beta, \beta_{max}, \lambda$ and $\epsilon=1e^{-6}$
		\REPEAT
		\STATE Update $\mathbf{\breve{S}}^{\sharp}$ by~\cref{solution_of_X} yields $\mathbf{\breve{S}}{\_}^{\sharp}$
            \STATE Update $\mathbf{\hat{S}}$ by~\cref{solution_for_S_hat} yields $\mathbf{\hat{S}}{\_}$
            \STATE Update $\mathbf{\tilde{S}}$ by~\cref{solution_for_S_tilde} yields $\mathbf{\tilde{S}}{\_}$
            \STATE Update $\mathbf{S}$ by~\cref{solution_for_S} yields $\mathbf{S}{\_}$
            \REPEAT
            \STATE Calculate the gradient by~\cref{eq:grad_for_admm_Q} and update $\mathbf{Q}$ by LM algorithm yields $\mathbf{Q}{\_}$
            \UNTIL \textbf{Convergence}
            \STATE Update $\mathbf{Y}_{1}$ by $\mathbf{Y}_{1} + \beta\left ( \mathbf{\breve{S}}{\_}^{\sharp} - g(\mathbf{\hat{S}}{\_}\mathbf{\Lambda}) \right)$
            \STATE Update $\mathbf{Y}_{2}$ by $\mathbf{Y}_{2} + \beta\left ( \mathbf{\hat{S}}{\_} - \mathbf{Q}{\_}\mathbf{\tilde{S}}{\_} \right)$
            \STATE Update $\mathbf{Y}_{3}$ by $\mathbf{Y}_{3} + \beta\left ( \mathbf{\tilde{S}}{\_} - \mathbf{R}_{p}\mathbf{{S}}{\_} \right)$
	    \STATE Update $\beta$ as $\min(\beta_{max}, \lambda\beta)$
            \UNTIL $\left \| \mathbf{S}{\_} - \mathbf{S} \right \|_{\infty}<\epsilon$
	\ENSURE  $\mathbf{S},\mathbf{Q}$
	\end{algorithmic}  
 
\end{algorithm}

\noindent\textbf{Solution for $\mathbf{S}$.}  The optimization model for the variable $\mathbf{{S}}$ can be expressed as follows:
\begin{equation}
    \begin{aligned}\label{S}
        \mathbf{{S}} & = \arg \min_{\mathbf{{S}}}~\frac{\mu_{1}}{2}\left \| \mathbf{W} - \mathbf{\Pi}\mathbf{S} \right \|^{2}_{F} \\
        & + \frac{\beta}{2}\left \| \mathbf{\tilde{S}} - \mathbf{R}_{p}\mathbf{{S}} \right \|_{F}^{2}+\left \langle \mathbf{Y}_{3}, \mathbf{\tilde{S}} - \mathbf{R}_{p}\mathbf{{S}} \right \rangle.
    \end{aligned}
\end{equation}
Calculating the derivative of model \eqref{S} and equating it to zero yields the closed-form solution of $\mathbf{{S}}$ as follows:
\begin{equation}
    \begin{aligned}\label{solution_for_S}
        (\frac{\mu_{1}}{\beta}\mathbf{\Pi}^{T}\mathbf{\Pi} & + \mathbf{I}_{3F})\mathbf{{S}} = \\
        & \frac{\mu_{1}}{\beta}\mathbf{\Pi}^{T}\mathbf{W} + \mathbf{R}_{p}^{T}\mathbf{\tilde{S}} + \frac{1}{\beta}\mathbf{R}_{p}^{T}\mathbf{Y}_{3}.
    \end{aligned}    
\end{equation}

\noindent\textbf{Solution for $\mathbf{Q}$.} We have already discussed how to solve the TPA module in~\cref{algo_for_spa}. But in model \eqref{lagrangian_formula}, the variables in $\Omega$ are coupled and need to be optimized alternatively, so the updating formula for $\mathbf{Q}$ needs adjustments. The optimization terms in model \eqref{lagrangian_formula} containing the optimization variable $\mathbf{Q}_{i},i=1,\cdots,F$ are:
\begin{equation}
    \begin{aligned}\label{Q_i}
        \mathbf{Q}_{i} & = \arg \min_{\mathbf{Q}_{i}}~\frac{\mu_{3}}{2}\sum_{i=1}^{F-1}\left \| \mathbf{Q}_{i}\mathbf{\tilde{S}}_{i} -\mathbf{Q}_{i+1}\mathbf{\tilde{S}}_{i+1} \right \|_{F}^{2} \\
        & + \frac{\beta}{2}\left \| \mathbf{\hat{S}} - \mathbf{Q}\mathbf{\tilde{S}} \right \|_{F}^{2}+\left \langle \mathbf{Y}_{2},\mathbf{\hat{S}} - \mathbf{Q}\mathbf{\tilde{S}} \right \rangle \\
        & = \arg \min_{\mathbf{Q}_{i}}~\frac{\mu_{3}}{2}\sum_{k=i-1}^{i}\sum_{j=1}^{P}{\left \| \mathbf{Q}_{k}\mathbf{\tilde{s}}_{k,j} - \mathbf{Q}_{k+1}\mathbf{\tilde{s}}_{k+1,j} \right \|}^{2}_{2} \\
        & + \frac{\beta}{2}\sum_{j=1}^{P}(\left \| \mathbf{\hat{s}}_{i,j} - \mathbf{Q}_{i}\mathbf{\tilde{s}}_{i,j} \right \|_{2}^{2} + \left \langle \mathbf{Y}_{2,j}^{i},\mathbf{\hat{s}}_{i,j} - \mathbf{Q}_{i}\mathbf{\tilde{s}}_{i,j} \right \rangle),
    \end{aligned}
\end{equation}
where $\mathbf{Y}_{2,j}^{i}$ is the rows $3i-2$ through $3i$ and $j$-th column of $\mathbf{Y}_{2}$. We denote $\mathbf{\hat{s}}_{i,j} - \mathbf{Q}_{i}\mathbf{\tilde{s}}_{i,j}$ as $\mathbf{\hat{r}}_{i,j}$ and compute the gradient of the Lie algebra $\phi_{i}$ corresponding to $\mathbf{Q}_{i}$ by imitating~\cref{eq:grad_for_spa}:
\begin{equation}
    \begin{aligned}\label{eq:grad_for_admm_Q}
     \mathbf{\hat{g}}_{i} & = \frac{\partial \frac{\mu_{3}}{2} \sum_{j=1}^{P}{(\left \| \mathbf{r}_{i,j}^{(1)} \right \|^{2}_{2}+\left \| \mathbf{r}_{i,j}^{(0)} \right \|^{2}_{2})}}{\partial \phi_{i}} \\
    & + \frac{\partial \sum_{j=1}^{P}{(\frac{\beta}{2}\left \| \mathbf{\hat{r}}_{i,j} \right \|^{2}_{2} + \left \langle  \mathbf{Y}_{2,j}^{i},\mathbf{\hat{r}}_{i,j} \right \rangle)}}{\partial \phi_{i}} \\ 
    & = \mu_{3} \sum_{j=1}^{P}{\left[ (\frac{\partial \mathbf{r}_{i,j}^{(1)}}{\partial \phi_{i}})^{T} \mathbf{r}_{i,j}^{(1)} + (\frac{\partial \mathbf{r}_{i,j}^{(0)}}{\partial \phi_{i}})^{T}\mathbf{r}_{i,j}^{(0)} \right]} \\
    & + \sum_{j=1}^{P}{ \left[{\beta}( \frac{\partial \mathbf{\hat{r}}_{i,j}}{\partial \phi_{i}})^{T} \mathbf{\hat{r}}_{i,j} + (\frac{\partial \mathbf{\hat{r}}_{i,j}}{\partial \phi_{i}})^{T}\mathbf{Y}_{2,j}^{i} \right ]} \\
     & = \sum_{j=1}^{P}{\left [ \left ( \mathbf{Q}_{i}\mathbf{\tilde{s}}_{i,j} \right )^{\wedge } \right ]^{T}}\left [\mu_{3} \left ( \mathbf{r}_{i,j}^{(1)} - \mathbf{r}_{i,j}^{(0)} \right ) + \beta \mathbf{\hat{r}}_{i,j} + \mathbf{Y}_{2,j}^{i}\right ].
\end{aligned}
\end{equation}
Therefore, we can still update $\mathbf{Q}_{i}$ using the TPA optimization algorithm in~\cref{algo_for_spa}, and only need to adjust the residual vector $\mathbf{r}_{i}$ to satisfy the descent direction.

After discussing the solution formulas for each variable, we give the complete optimization~\cref{alg2}. Since no closed-form solution for updating $\mathbf{Q}$ exists, another iterative optimization must be embedded in the ADMM algorithm. In experiments, we found that only $1$ to $10$ iterations are needed to update $\mathbf{Q}$ well.

\begin{figure*}
  \centering
  \includegraphics[height=16cm,width=1.0\linewidth]{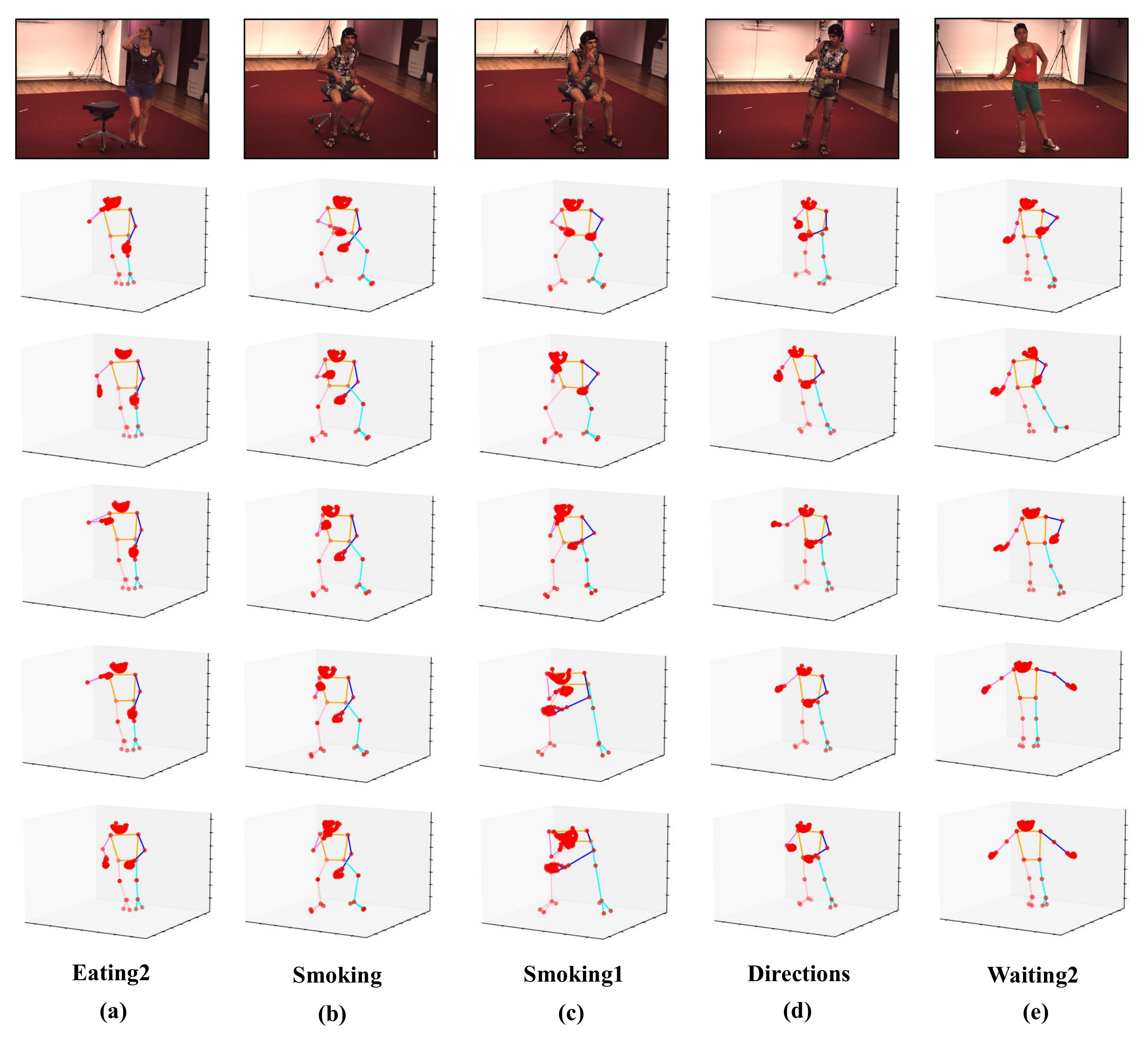}
  \caption{Visualization of the 3D shapes in the H3WB NRSfM dataset. The first row shows the pictures in the original H36M dataset, and the other rows display the GT 3D shapes in each sequence.}
  \label{fig:h3wb_dataset}
\end{figure*}

\subsection{Model with Occlusion}\label{model_occl}
In real-world scenes, the captured images are often obscured and it is difficult to observe all the keypoints in each frame. Assume $\mathbf{o}_{i}\!\in\!\mathbb{R}^{1\times P},i=1,\!\cdots\!,F$ is the mask vectors and $o_{i,j}\!=\!1$ if $j$-th point in the $i$-th frame is visible, otherwise $0$.
Then to solve the occlusion problem, we introduce the mask matrix $\mathbf{O}\!\in\!\mathbb{R}^{2F \times P}$ and correct the data term constraint $\mathcal{F}(\cdot)$ as follows:
\vspace{-0.25em}
\begin{equation}
    \frac{\mu_{1}}{2}\left \| \mathbf{O}  \odot \left ( \mathbf{W} - \mathbf{\Pi}\mathbf{S} \right) \right \|_{F}^{2},
\vspace{-0.25em}
\end{equation}
where $\mathbf{O} = \left [ \mathbf{1}_{2} \otimes \mathbf{o}_{1}; \cdots;\mathbf{1}_{2} \otimes \mathbf{o}_{F} \right ] \in \mathbb{R}^{2F \times P}$. Since the occlusion is different for each frame, centralization for the observation matrix $\mathbf{W}$ does not guarantee that the translation between shapes is eliminated. Therefore re-centering of the shapes is required before alignment using the TPA module, which only requires modification of $\mathbf{\tilde{S}}$:
\vspace{-0.25em}
\begin{equation}
\mathbf{\tilde{S}}_{i}=\mathbf{R}_{pi}\mathbf{{S}}_{i}\mathbf{T},i=1,\cdots,F,
\vspace{-0.25em}
\end{equation}
where the definition of $\mathbf{T}=\mathbf{I}-\frac{1}{P}\mathbf{1}\mathbf{1}^{T}$ is translation removal matrix. The completed model after adding occlusion is as follows:
\vspace{-0.5em}
\begin{equation}
\begin{aligned}\label{occl_model}
     \min_{\mathbf{S},\mathbf{Q}} & \frac{\mu_{1}}{2}\left \| \mathbf{O}  \odot \left ( \mathbf{W} - \mathbf{\Pi}\mathbf{S} \right) \right \|^{2}_{F} + \mu_{2}\left \| \mathbf{\breve{S}}^{\sharp} \right \|_{*}+ 
      \\ & \frac{\mu_{3}}{2}\sum_{i=1}^{F-1}\left \| \mathbf{Q}_{i}\mathbf{\tilde{S}}_{i} -\mathbf{Q}_{i+1}\mathbf{\tilde{S}}_{i+1} \right \|_{F}^{2} \\
    s.t. & \left\{\begin{array}{l}
\mathbf{\breve{S}}^{\sharp} = g(\mathbf{\hat{S}}\mathbf{\Lambda} )\\ 
\mathbf{\hat{S}}_{i}=\mathbf{Q}_{i}\mathbf{\tilde{S}}_{i},i=1,\cdots,F\\ 
\mathbf{\tilde{S}}_{i}=\mathbf{R}_{pi}\mathbf{{S}}_{i}\mathbf{T},i=1,\cdots,F
\end{array}\right.
\end{aligned}
 \raisetag{6\baselineskip}
\end{equation}
The update formulas for variables $\Omega$ can be obtained by imitating the above solution procedure. Since the improvement of the model is only related to the optimization variable $\mathbf{\tilde{S}}, \mathbf{S}$, we only need to adjust their update formulas.

\noindent \textbf{Solution for $\mathbf{\tilde{S}}$ under occlusion.} $\mathbf{S}$ in model \eqref{occl_model} needs to be centralized and then transformed by rotation $\mathbf{R}_{pi}$ to get $\mathbf{\tilde{S}}$. Thus we only need to replace $\mathbf{S}$ in~\cref{solution_for_S_tilde} with $\mathbf{ST}$, \ie:
\begin{equation}
    \begin{aligned}\label{solution_for_S_tilde_occl}
        (\frac{\mu_{3}}{\beta}\mathbf{Q}^{T}\mathbf{H}^{T}&\mathbf{H}\mathbf{Q} + 2\mathbf{I}_{3F})\mathbf{\tilde{S}} = \\
        & \mathbf{Q}^{T}\mathbf{\hat{S}} + \frac{1}{\beta}\mathbf{Q}^{T}\mathbf{Y}_{2} + \mathbf{R}_{p}\mathbf{{ST}} - \frac{1}{\beta}\mathbf{Y}_{3}.
    \end{aligned}    
\end{equation}
\noindent \textbf{Solution for $\mathbf{S}$ under occlusion.} We solve the 3D shape $\mathbf{S}_{i}$ frame-by-frame and the optimization model is as follows:
\begin{equation}
    \begin{aligned}\label{S_occl}
        \mathbf{{S}}_{i} & = \arg \min_{\mathbf{{S}}_{i}}~\frac{\mu_{1}}{2}\left \| \mathbf{O}_{i}  \odot \left ( \mathbf{W}_{i} - \mathbf{\Pi}_{i}\mathbf{S}_{i} \right) \right \|^{2}_{F} \\
        & + \frac{\beta}{2}\left \| \mathbf{\tilde{S}}_{i} - \mathbf{R}_{pi}\mathbf{{S}}_{i}\mathbf{T} \right \|_{F}^{2}+\left \langle \mathbf{Y}_{3}^{i}, \mathbf{\tilde{S}}_{i} - \mathbf{R}_{pi}\mathbf{{S}}_{i}\mathbf{T} \right \rangle,
    \end{aligned}
\end{equation}
where $\mathbf{O}_{i} = \mathbf{1}_{2}\otimes \mathbf{o}_{i}$. We define $\mathbf{M}_{i}=diag(\mathbf{o}_{i}) \in \mathbb{R}^{P\times P}$, then model \eqref{S_occl} can be rewritten as:
\begin{equation}
    \begin{aligned}\label{S_occl2}
        \mathbf{{S}}_{i} & = \arg \min_{\mathbf{{S}}_{i}}~\frac{\mu_{1}}{2}\left \|  \left ( \mathbf{W}_{i} - \mathbf{\Pi}_{i}\mathbf{S}_{i} \right)\mathbf{M}_{i} \right \|^{2}_{F} \\
        & + \frac{\beta}{2}\left \| \mathbf{\tilde{S}}_{i} - \mathbf{R}_{pi}\mathbf{{S}}_{i}\mathbf{T} \right \|_{F}^{2}+\left \langle \mathbf{Y}_{3}^{i}, \mathbf{\tilde{S}}_{i} - \mathbf{R}_{pi}\mathbf{{S}}_{i}\mathbf{T} \right \rangle.
    \end{aligned}
\end{equation}
Calculating the derivative of model \eqref{S_occl2} and equating it to zero yields the solution of $\mathbf{{S}}_{i}$ as follows:
\begin{equation}
    \begin{aligned}\label{solution_for_S_occ}
        \frac{\mu_{1}}{\beta}\mathbf{\Pi}_{i}^{T}\mathbf{\Pi}_{i}&\mathbf{{S}}_{i}\mathbf{M}_{i}^{2} + \mathbf{{S}}_{i}\mathbf{T}^{2} = \\
        & \frac{\mu_{1}}{\beta}\mathbf{\Pi}_{i}^{T}\mathbf{W}_{i}\mathbf{M}_{i}^{2} + \mathbf{R}_{pi}^{T}\mathbf{\tilde{S}}_{i}\mathbf{T} + \frac{1}{\beta}\mathbf{R}_{pi}^{T}\mathbf{Y}_{3}^{i}\mathbf{T}.
    \end{aligned}    
\end{equation}
In order to calculate the closed-form solution of $\mathbf{S}_{i}$, we use the property $vec(\mathbf{AXB})=(\mathbf{B}^{T}\otimes\mathbf{A})vec(\mathbf{X})$ to equivalently represent~\cref{solution_for_S_occ} as:
\begin{equation}
    \begin{aligned}\label{solution_for_S_occ2}
         &\left [\frac{\mu_{1}}{\beta}(\mathbf{M}_{i}^{2}\otimes \mathbf{\Pi}_{i}^{T}\mathbf{\Pi}_{i}) + (\mathbf{T}^{2}\otimes \mathbf{I}_{3})\right]vec(\mathbf{{S}}_{i}) = \\
        &~~~~~vec\left(\frac{\mu_{1}}{\beta}\mathbf{\Pi}_{i}^{T}\mathbf{W}_{i}\mathbf{M}_{i}^{2} + \mathbf{R}_{pi}^{T}\mathbf{\tilde{S}}_{i}\mathbf{T} + \frac{1}{\beta}\mathbf{R}_{pi}^{T}\mathbf{Y}_{3}^{i}\mathbf{T}\right).
    \end{aligned}    
\end{equation}
The update formulas for the other optimization variables in Model \eqref{occl_model} remain unchanged.

\begin{figure}[t]
  \centering
  \includegraphics[width=1.0\linewidth]{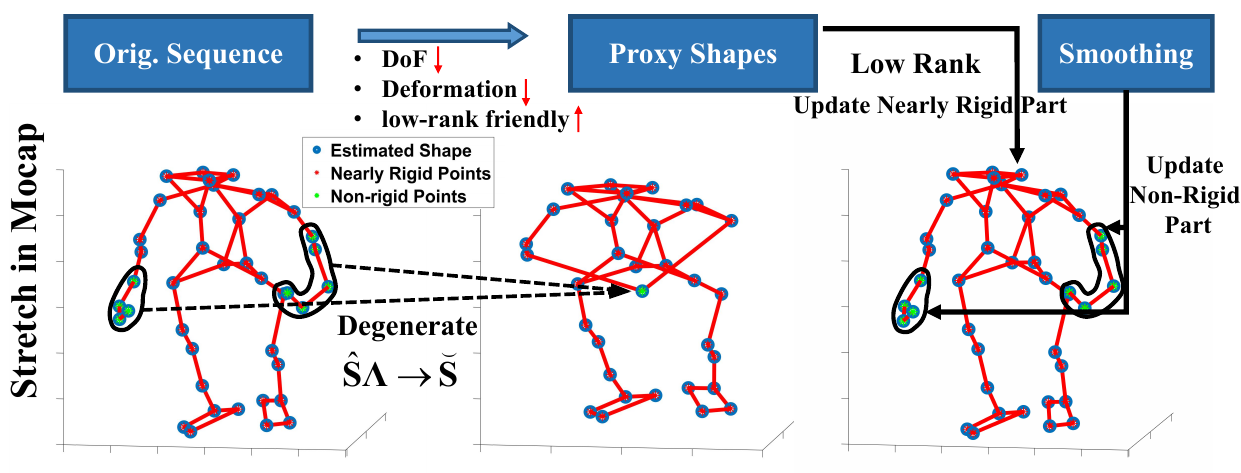}
  \caption{Illustration of the proxy shape construction and shape update principle.}
  \vspace{-0.2em}
   \label{fig:swnn_intro}
\end{figure}

\section{Supplement to Spatially-variant Modeling}\label{Spatial_variant_Modeling}
In this section, We provide an additional explanation of the construction of the proxy shape and how it plays a role in optimization, refer to~\cref{fig:swnn_intro}.

\noindent\textbf{1) Definition of proxy shape.} We divide the non-rigid object into two regions with different deformation degrees through frequency domain analysis of 3D trajectories. Then by spatial weighting, we merge the single super point degenerated from ``Non-Rigid'' part with ``Nearly Rigid'' points to form the \textbf{proxy shapes}, \ie, $\mathbf{\breve{S}}\!=\!\mathbf{\hat{S}}\mathbf{\Lambda}$. 
The weight matrix $\mathbf{\Lambda}$ defined by the feature mapping $\phi(\cdot)$ (Eq.~(12) in them main text) is rank-deficient (when $\alpha_{r}<1$), so the resultant proxy sequence $\mathbf{\breve{S}}$ has fewer degrees of freedom than $\mathbf{\hat{S}}$ and more satisfies the low-rank constraint.

\noindent\textbf{2) How SWNN works.} We enforce the low-rank constraint on proxy shapes and update new shapes by~\cref{solution_for_S_hat}. 
The new $\mathbf{\hat{S}}$ is mainly composed of $\mathbf{\breve{S}}$ after low-rank regularization and $\mathbf{\tilde{S}}$ after smoothing regularization.
Since $\mathbf{\Lambda}$ is rank-deficient, the low-rank constrained $\mathbf{\breve{S}}$ after inverse transformation only retains the structural information of nearly rigid part. Therefore, $\mathbf{\hat{S}}$ updates nearly rigid part mainly through low-rank, and non-rigid part is fine-tuned by $\mathbf{\tilde{S}}$.

\begin{figure}
  \centering
  \includegraphics[width=1.0\linewidth]{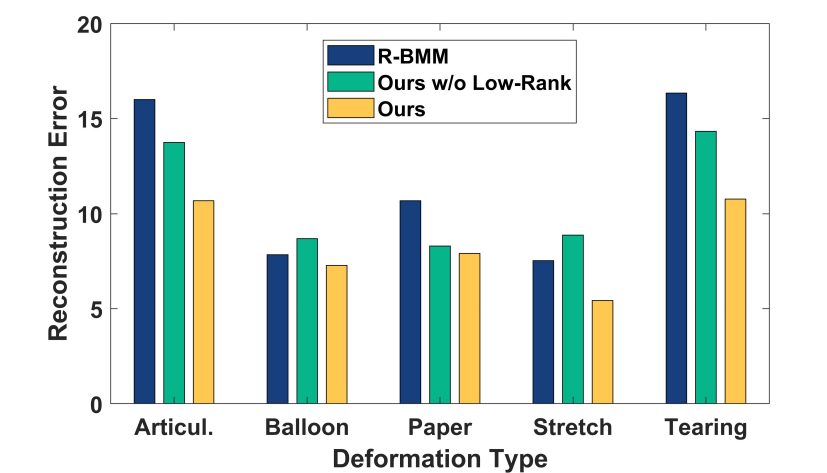}
  \caption{Comparison of the reconstruction errors of our model with methods using only low-rank or smoothing constraints on the NRSfM Challenge dataset.}\vspace{-0.4cm}
  \label{fig:pure_smooth}
\end{figure}

Our method combines low-rank and smoothing constraints through spatial weighting, which improves the accuracy of the reconstruction by avoiding the over-penalization of localized drastic deformations with low-rank constraint. To verify the validity of the combination, we designed the comparison experiment presented in~\cref{fig:pure_smooth}. We compare the reconstruction error of~\cref{lagrangian_formula} with methods using only low-rank or smoothing constraints on the NRSfM Challenge dataset. For the low-rank-only method, we compared with R-BMM~\cite{kumar2020non}, which is an improvement on the ``prior-free" classical method BMM~\cite{dai2014simple}. For the approach using only the smooth prior, we removed the low-rank constraint from~\cref{lagrangian_formula} and compared with it, \ie:
\begin{equation}
\resizebox{0.45\textwidth}{!}{$
    \begin{aligned}
     \min_{\mathbf{\tilde{S}}, \mathbf{{S}}, \mathbf{Q}} ~ \mathcal{L} & =  \frac{\mu_{1}}{2}\left \| \mathbf{W}\!-\!\mathbf{\Pi}\mathbf{S} \right \|^{2}_{F}  + \frac{\mu_{2}}{2}\sum_{i=1}^{F-1}\left \| \mathbf{Q}_{i}\mathbf{\tilde{S}}_{i}\!-\!\mathbf{Q}_{i+1}\mathbf{\tilde{S}}_{i+1} \right \|_{F}^{2} \\
     & + \frac{\beta}{2}\left \| \mathbf{\tilde{S}} - \mathbf{R}_{p}\mathbf{{S}} \right \|_{F}^{2}+\left \langle \mathbf{Y}_{1}, \mathbf{\tilde{S}} - \mathbf{R}_{p}\mathbf{{S}} \right \rangle .
 \end{aligned}$}
\end{equation}
As shown in~\cref{fig:pure_smooth}, the reconstruction error of our method is lower than that of the low-rank-only/smooth-only method on all types of deformation, and there is a significant improvement on Articul., Stretch and Tearing. 

Our approach can effectively couple low-rank and smoothing constraints to improve the stability of the algorithm over different types of deformations.
Moreover, statistical prior methods tend to be sensitive to the statistical properties of the data, which are also related to the complexity of the corresponding object deformation. In other words, it is not enough to only mine the smooth prior to complement the low-rank constraints, so finding constraints that are more general and insensitive to the statistical properties of the data is a feasible direction for improvement.

\begin{figure}
  \centering
  \includegraphics[width=1.0\linewidth]{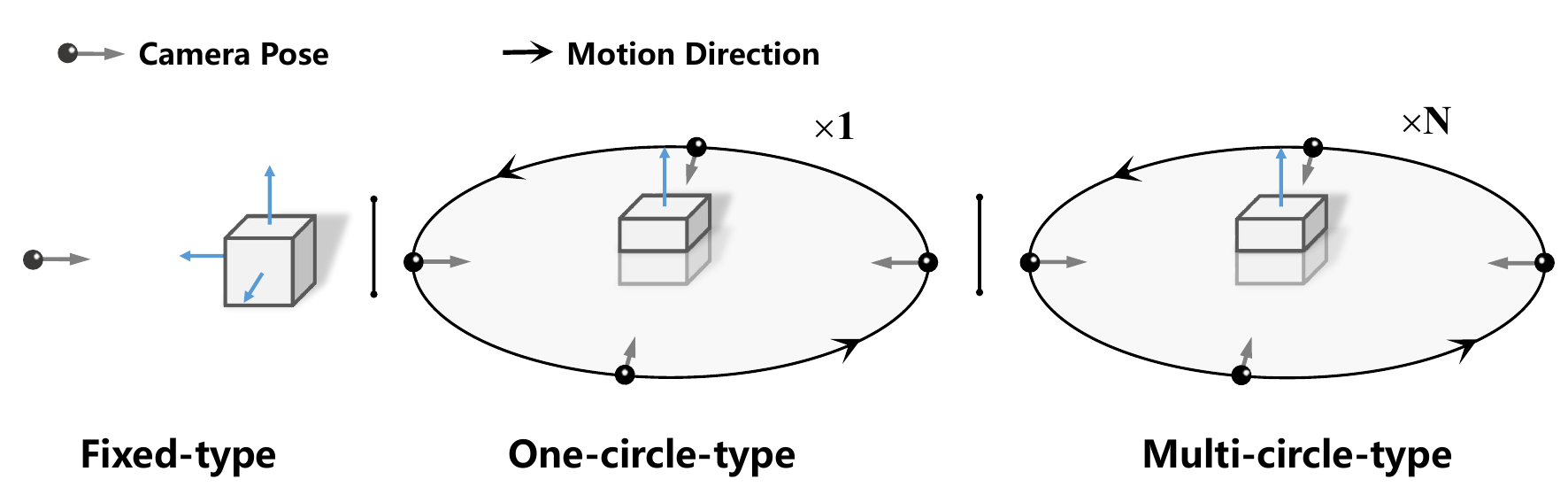}
  \caption{Three motion types for H3WB Dataset generation.}
   \label{fig:h3wb_motion}
\end{figure}

\begin{figure*}
  \centering
  \includegraphics[width=1.0\linewidth]{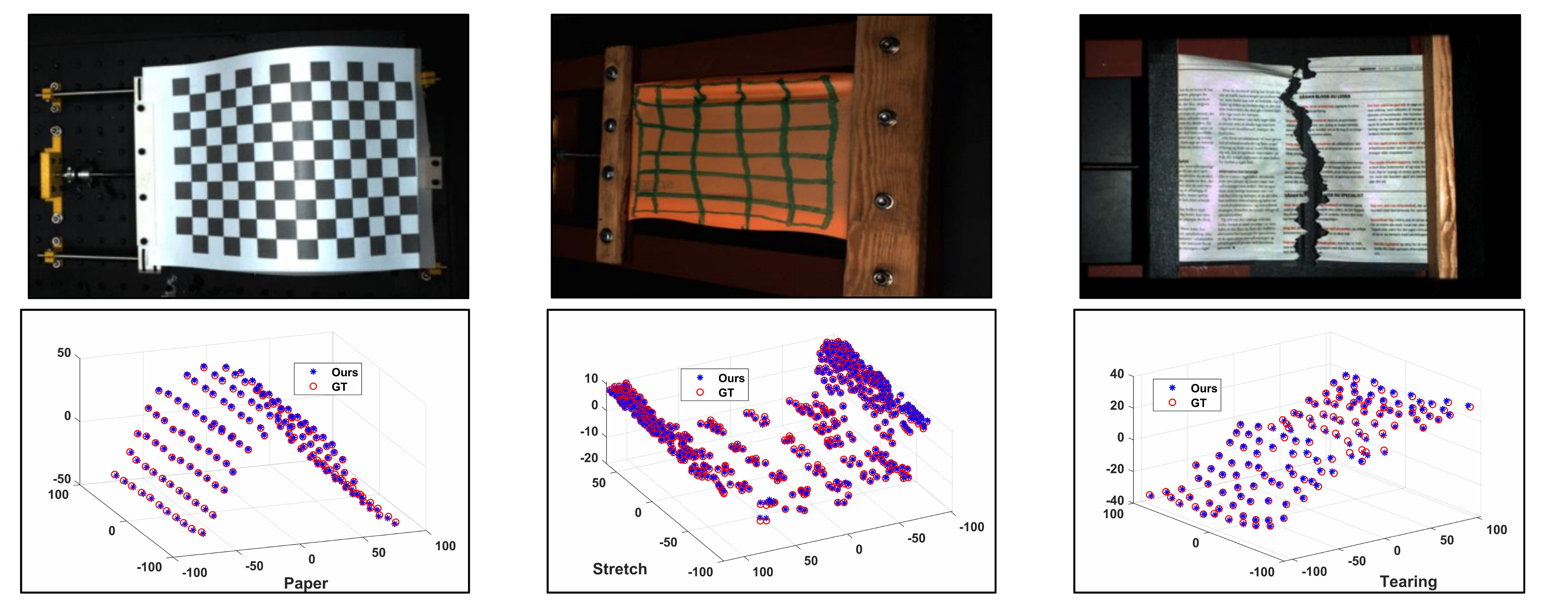}
  \caption{Qualitative Results on the NRSfM Challenge Benchmark. The first row shows the images in the dataset, and the second row shows the reconstruction results of our method compared to the GT.}
  \label{fig:supp_chal_vis}
\end{figure*}

\begin{figure*}
  \centering
  \includegraphics[width=1.0\linewidth]{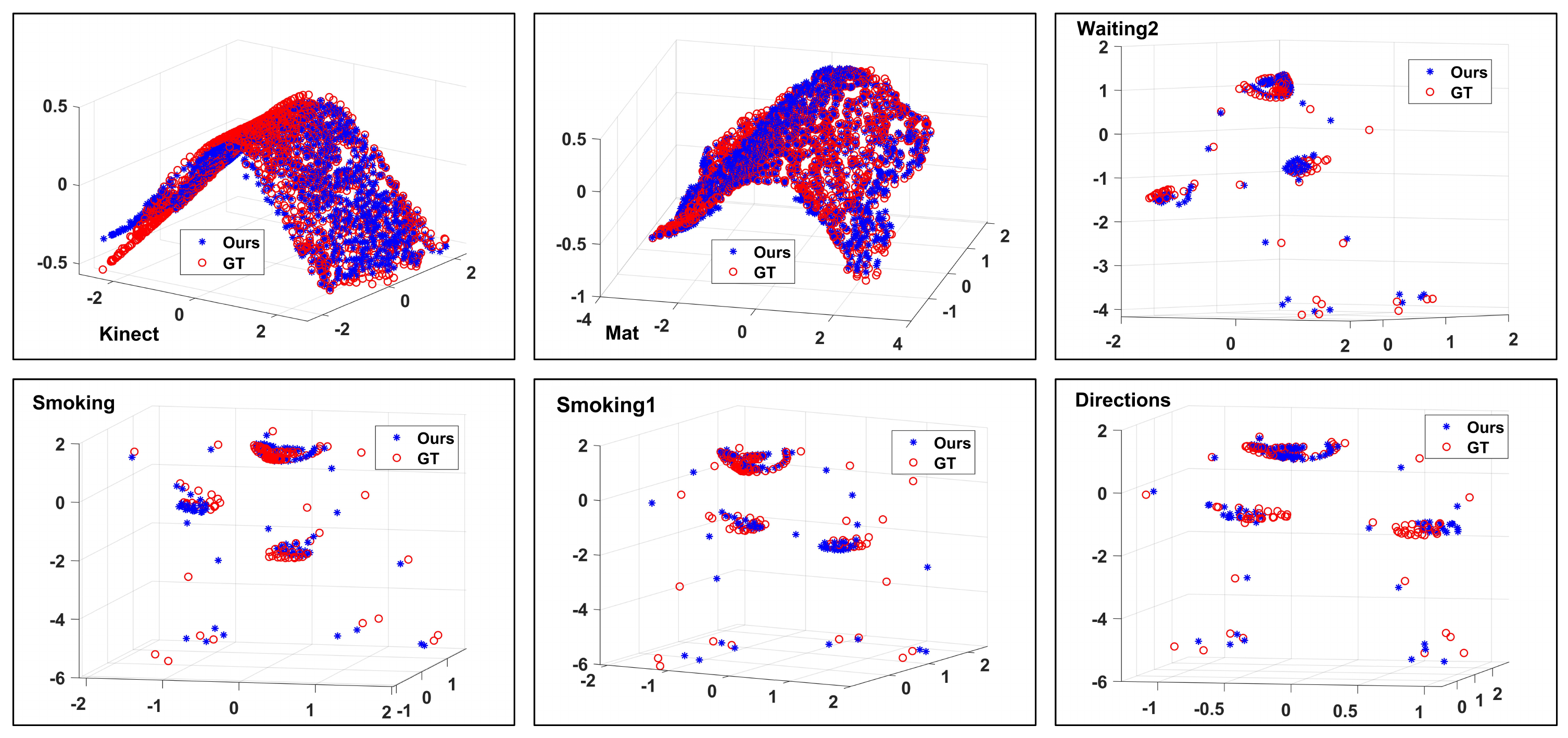}
  \caption{Qualitative Results on the Semi-dense and H3WB dataset (Fixed-type). The reconstruction results of our method on the Semi-dense dataset are closer to the GT shapes, while the H3WB dataset is very challenging and cannot yet be accurately reconstructed.}
  \label{fig:supp_dense2h3wb_vis}
\end{figure*}

\begin{figure}[t]
  \centering
  \includegraphics[width=1.0\linewidth]{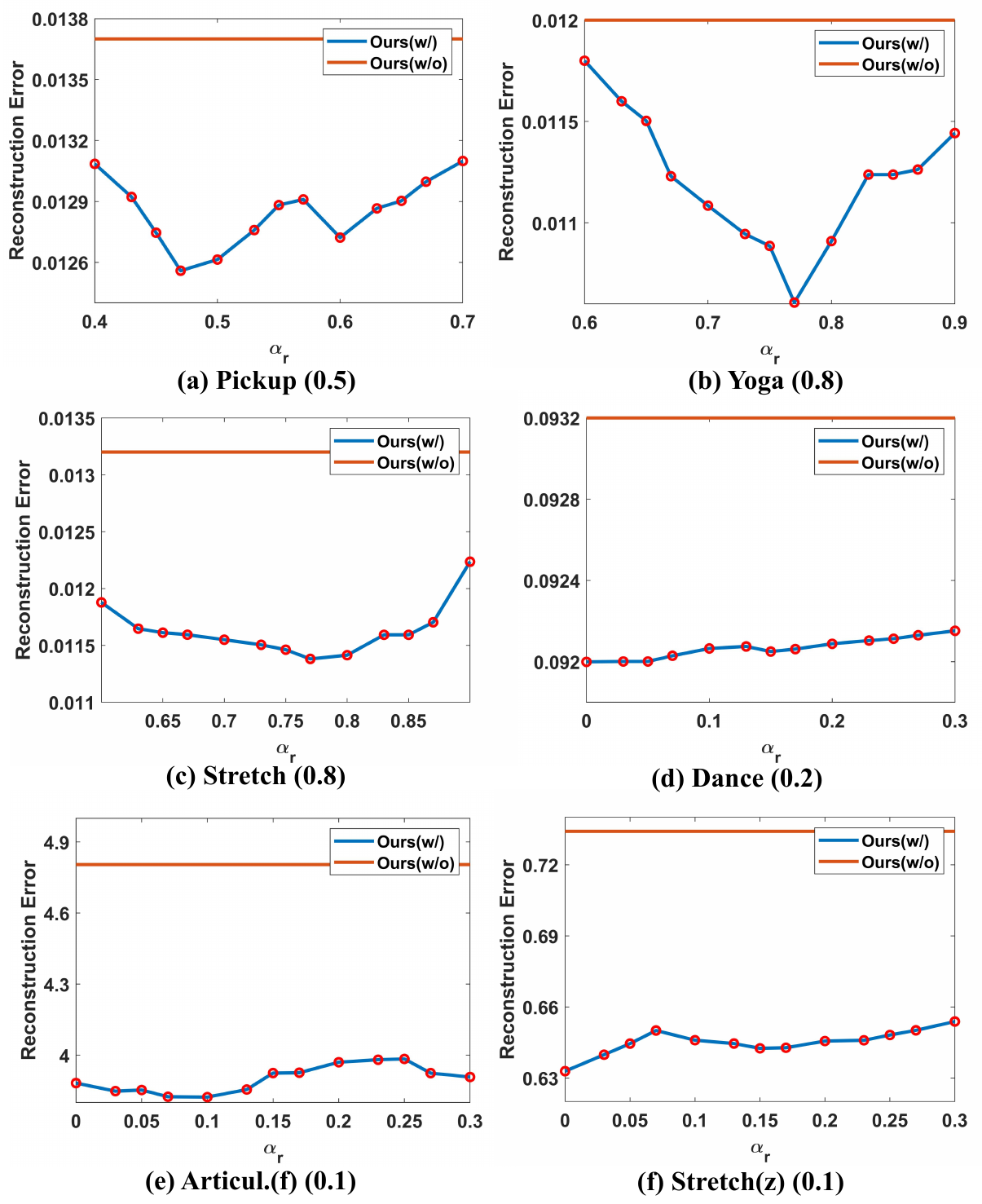}
  \caption{Stability test for hyperparameter $\alpha_{r}$ on different sequences. Straight lines indicate the results without the SWNN module and broken lines indicate the reconstruction errors for different $\alpha_{r}$ settings. To the right of the sequence names are the $\alpha_r$-values corresponding to the results reported in the main text.}\vspace{-0.4cm}
  \label{fig:alpha_r}
\end{figure}

\section{Additional Experiments}
In this section, we first add the details of dataset processing and algorithm implementation and then report additional experiment results.
\subsection{Implementation Details}
For parameters in ADMM~\cref{alg2}, we refer to~\cite{kumar2022organic} to initialize $\beta\!=\!1e^{-4}$, $\beta_{max}\!=\!1e^{10}$, $\mathbf{\lambda}\!=\!1.1$ and set $\{ \mathbf{Y}_{n} \}_{n=1}^{3}$ to zero matrices.
The weights $\mu_{1}, \mu_{2}, \mu_{3}$ of the optimization objective~\eqref{lagrangian_formula} are set to $1, 0.1, 0.1$ by default. We can adjust the sizes of $\mu_{2}, \mu_{3}$ according to the shape sequences' low-rank and smoothing properties. In addition, we found if the shape sequence does not satisfy the low-rank property well, increasing the weight $\mu_{1}$ of the reprojection term can effectively improve the reconstruction results, \eg, $\mu_{1}\!=\!1e^{1}$ generally on the NRSfM Challenge dataset and $\mu_{1}\!=\!1e^{2}$ on kinect and rug in the Semi-dense dataset. For the estimation of proxy shape, we adjust the ratio of nearly rigid points $\alpha_{r}$ between $0$ to $1$ depending on the deformation characteristics of the object. $\delta_{r}$ is set to $\frac{1}{3}$ on Mocap and NRSfM Challenge datasets and $0$ on Semi-dense and H3WB datasets.

As described in~\cref{Spatial_variant_Modeling}, $\alpha_{r}$ determines the optimization approach used for different regions of the 3D structure, so $\alpha_{r}$ should depend on the specific characteristics of the object's deformation, \eg, spatial coherence, continuity, \etc. We adopted a testing interval of $0.2$ for parameter selection initially, followed by a finer search around promising results using a step size of $0.1$. This granularity proved sufficient for achieving good outcomes, eliminating the need for further refinement in the search steps.
The stability test for our method on $\alpha_{r}$ is illustrated in~\cref{fig:alpha_r}. In most cases, the algorithm is stable with respect to $\alpha_{r}$ once a rough selection interval has been determined, and thus careful screening of $\alpha_{r}$ is not necessary. When we have a large amount of data, we can segment the keypoints by learning a certain distribution without setting hyperparameters, like~\cite{del2006non}.

\begin{figure}
  \centering
  \begin{subfigure}{0.98\linewidth}
    \includegraphics[width=1.0\linewidth]{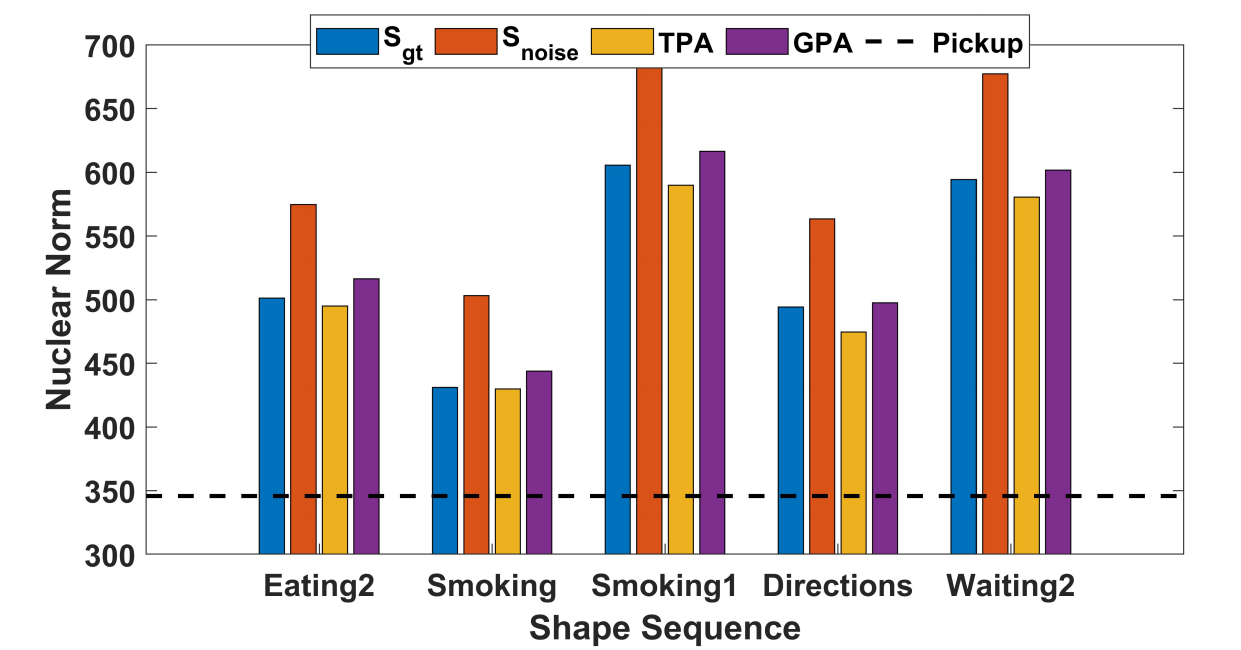}
    \caption{Low-rank Property of Shape Sequences in H3WB Dataset}
    \label{fig:h3wb_lr}
  \end{subfigure}
  \hfill
  \begin{subfigure}{0.98\linewidth}
    \includegraphics[width=1.0\linewidth]{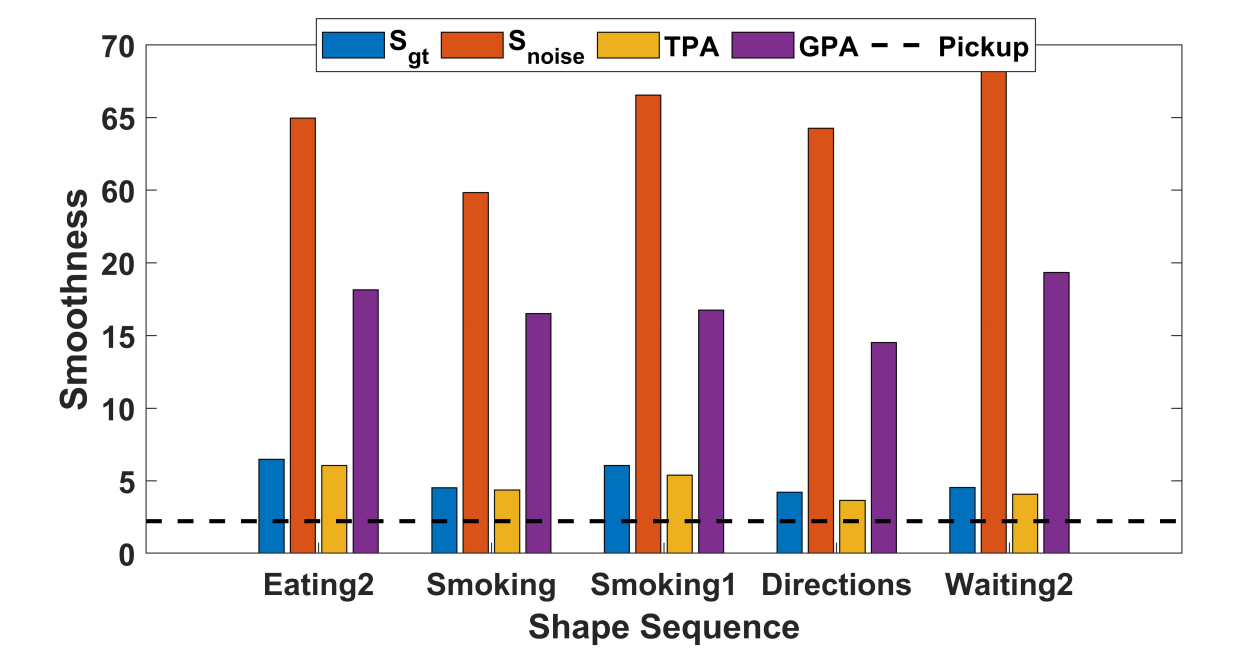}
    \caption{Smoothing Property of Shape Sequences in H3WB Dataset}
    \label{fig:h3wb_smth}
  \end{subfigure}
  \caption{Analyzing the low-rank and smoothing properties of shape sequences in the H3WB dataset. The black dotted line represents the values of the metrics for the pickup sequence in the Mocap dataset. The comparison reveals that the sequences in the H3WB dataset have a greater magnitude of motion and are more difficult to recover using low-rank and smoothing constraints.}
  \vspace{-1.0em} 
  \label{fig:analysis_for_h3wb}
\end{figure}

\subsection{H3WB Dataset Processing}
The H3WB dataset~\cite{zhu2023h3wb} is an entire human body 3D dataset extended from the H36M dataset~\cite{ionescu2013human3}. However, this dataset provides 2D annotations and their corresponding 3D structures frame by frame rather than in a sequence. We screened with the criterion of being as contiguous as possible and obtained five sequences from \textbf{S1 Eating2}, \textbf{S6 Smoking}, \textbf{S6 Smoking1}, \textbf{S6 Directions}, and \textbf{S7 Waiting2}. Since the motion amplitude of these sequences is vast, we remove the frames with significant mutations and use spline functions to interpolate the remaining parts to obtain five more realistic human action sequences: Eating2 (185 frames), Smoking (180), Smoking1 (265), Directions (245), and Waiting2 (335). To obtain the coordinates of the 2D keypoints under the orthogonal projection model, we set up three camera motion types: \textbf{Fixed-type}, \textbf{One-circle-type}, \textbf{Multi-circle-type} (as in~\cref{fig:h3wb_motion}). 1) For Fixed-type, the camera is fixed to a certain viewpoint; 2) For One-circle-type, assume that the video has $F$ frames and the camera rotates at speed $\frac{1}{F}^{\circ}/frame$ around a direction vertical to the ground. In other words, the camera orbits the object exactly one time during its deformation; 3) For Multi-circle-type, we keep the angular velocity of the camera motion fixed at $5^{\circ}/frame$ in reference to~\cite{akhter2008nonrigid}, \ie, the camera goes around the object every 120 frames. Theoretically, the difficulty of sequence reconstruction is ranked Fixed-type$>$One-circle-type$>$Multi-circle-type.

We show a partial 3D shape sequences of the H3WB NRSfM dataset in~\cref{fig:h3wb_dataset}. We used the experimental setup in~\cref{tpa_exp_setting} to test the low-rank and smoothing properties of the H3WB dataset, and the results are displayed in~\cref{fig:analysis_for_h3wb}. The sequences in the H3WB dataset generally have higher nuclear norm and first-order smoothing errors compared to the pickup sequence in Mocap. Recovering the 3D shapes in the H3WB dataset using low-rank and smoothing constraints is more complicated. However, it is worth noting that the TPA-aligned sequences have better low-rank and smoothing properties than the GT sequences, which somewhat guarantees the validity of the low-rank regularization.

\subsection{Additional Experiments on Missing Data}
In reality, the movement of an object often leads to occlusion of different regions, resulting in missing 2D observations obtained from camera shots. Therefore, the stability of the algorithm on the missing dataset is significant. We show some tests on missing data in the main text, and we will add more quantitative results in this section.

\begin{table*}[t]
\centering
\scriptsize
\caption{\centering Comparison of reconstruction errors on missing data in Mocap and H3WB datasets. ' - ' indicates that test results are unavailable. Ours(w/) and Ours(w/o) represent the test results with and without occlusion respectively.}
\label{table_occl}
\vspace{-0.2cm}
\resizebox{0.9\linewidth}{!}{
\begin{tabular}{c|ccc|ccccc}
\hline
Method    & Shark          & Face         & Walking           & Eating2        & Smoking1          & Directions                    & Smoking     & Waiting2   \\ \hline
CSF2\cite{gotardo2011non}  & 0.0653 & 0.0412 & 0.1033 &	0.2556 & 0.3659 & 0.3025 & 0.4058 & 0.1862
 \\ 
PND\cite{lee2013procrustean} & 0.0166 & 0.0177 &	\textbf{0.0469} & 0.2167 & 0.2954 & 0.3054 & 0.4863 & 0.2256
   \\ 
PMP\cite{Lee_2014_CVPR} & \textbf{0.0116} & 0.0174 & 0.0507	&-	&-	&-	&-	&-  \\ 
Ours(w/) & 0.0268	& \textbf{0.0154} & 0.0796	&\textbf{0.1657} & \textbf{0.1971} & \textbf{0.2858}	& \textbf{0.1804}	& \textbf{0.1113}
   \\ \hline
Ours(w/o) & 0.0258	& {0.0144} & 0.0710	&{0.1767} & {0.2040} & {0.2688}	& {0.1693}	& {0.1097}
   \\ \hline
\end{tabular}}
\vspace{-1.0em}
\end{table*}

We use the model in~\cref{model_occl} and follow the settings in~\cite{lee2013procrustean} to randomly add masks (occlusion rate 30\%) to the 2D observation matrix $\mathbf{W}$. Before reconstruction, we first solve the low-rank approximation of the observation matrix to complement it~\cite{cabral2013unifying}, which is important for the initialization of the camera motion. \cref{table_occl} shows the test results on the remaining sequences of the Mocap dataset and the H3WB dataset. Our method achieves the best performance on sequences other than Shark and Walking. The experimental results show that the reconstruction errors on data with and without occlusion display consistency. And the comparison between them indicates that our method still possesses stability under random occlusion settings (around 30\% occlusion).

\begin{figure}
  \centering
  \begin{subfigure}{0.49\linewidth}
    \includegraphics[width=1.0\linewidth]{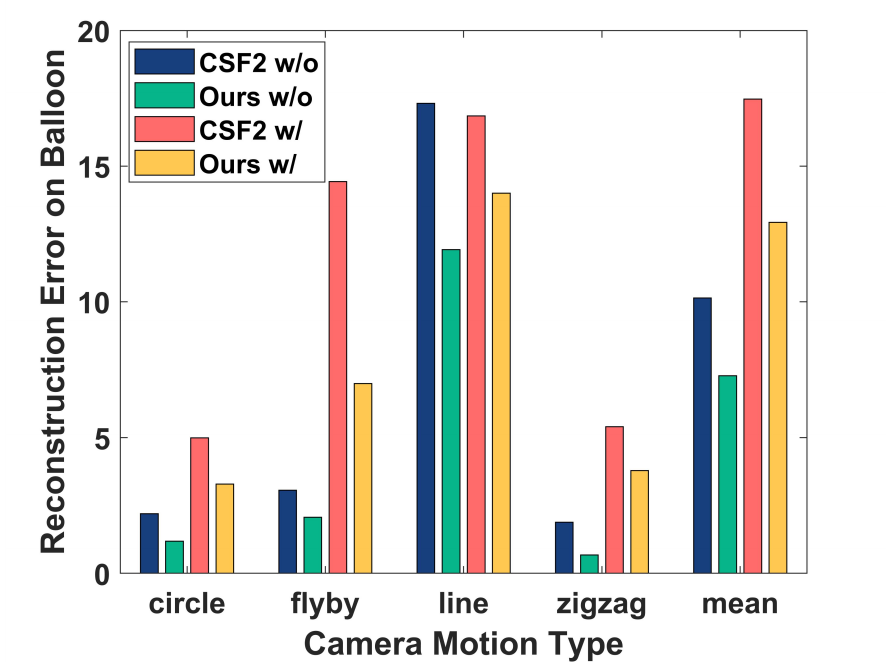}
    \caption{$e_{3d}$ on Seq. Balloon}
    \label{fig:balloon_occl}
  \end{subfigure}
  \hfill
  \begin{subfigure}{0.49\linewidth}
    \includegraphics[width=1.0\linewidth]{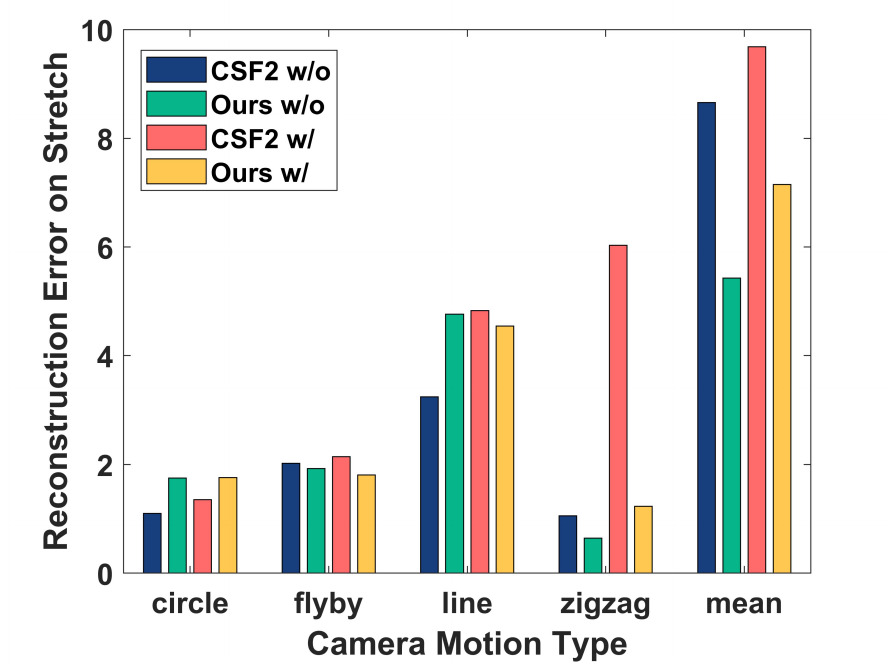}
    \caption{$e_{3d}$ on Seq. Stretch}
    \label{fig:stretch_occl}
  \end{subfigure}
  \caption{Reconstruction error on missing data in NRSfM Challenge Dataset. For a better comparison, the figure shows the test results under multiple camera motion types and compares the reconstruction errors without missing data.}\vspace{-1.0em}
  \label{fig:real_occl}
\end{figure}

Simulating occlusion by randomly adding masks is friendly to the recovery of observation matrix, whereas the occlusion scenario in reality tends to be more complex. We tested on the real occlusion data provided by the NRSfM Challenge Dataset, and the results are shown in~\cref{fig:real_occl}. The figure shows the reconstruction error comparison with the CSF2~\cite{gotardo2011computing} on two sequences Balloon and Stretch. Our approach is superior in terms of mean performance, regardless of whether the data is occluded or not. In addition, the occlusion rate is the main factor affecting the performance of the algorithm. The average occlusion rate under all camera motion types for Balloon is 38\%, while for Stretch it is 13\%. As a result the method's accuracy degradation is more obvious on Balloon. Apart from the occlusion rate, the accuracy of our method also relies on the results of matrix completion. Matrix completion based on the low-rank assumption tends to fail for some special occlusion scenarios, \eg, the object is completely invisible at some moments (such as Articulated/tricky in NRSfM Challenge Dataset, the object is completely occluded in the first 35 frames) and excessive occlusion (such as Tearing/tricky with the occlusion rate of 56\%). Searching for more robust matrix-completion algorithms or updating the observation matrix in iterations~\cite{paladini2009factorization} are potential solutions.

\subsection{Additional Qualitative Results}
In this section, we provide more visualizations of the 3D reconstruction results of our method. \cref{fig:supp_chal_vis} shows additional qualitative results on the NRSfM Challenge dataset, and~\cref{fig:supp_dense2h3wb_vis} illustrates 3D reconstruction results compared with GT on the Semi-dense and H3WB datasets (Fixed-type).


\end{document}